\documentclass[11pt]{article}
\usepackage{nodalida2025}
\usepackage{times}
\PassOptionsToPackage{hyphens}{url}
\usepackage{latexsym}
\usepackage{rotating} 
\usepackage{array}
\usepackage{amsmath}
\usepackage{marginnote}
\usepackage{colortbl}  

\aclfinalcopy

\usepackage{microtype}
\usepackage[utf8]{inputenc} 
\usepackage[T1]{fontenc}
\usepackage[table,xcdraw]{xcolor}
\definecolor{darkblue}{rgb}{0, 0, 0.5}
\usepackage[colorlinks=true, citecolor=darkblue, linkcolor=darkblue, urlcolor=darkblue]{hyperref}
\usepackage{url}            
\usepackage{booktabs}       
\usepackage{amsfonts}       
\usepackage{nicefrac}       
\usepackage{inconsolata}
\usepackage{enumitem}
\usepackage[frozencache,cachedir=.]{minted}
\usepackage{graphicx}
\setlist[itemize]{left=0.0cm}
\setlist[enumerate]{left=0.0cm}
\usepackage{comment}
\usepackage[nameinlink,capitalize,noabbrev]{cleveref}
\usepackage[skip=8pt]{caption}
\usepackage{tikz}
\DeclareRobustCommand*\circled[1]{\tikz[baseline=(char.base)]{
            \node[shape=circle,draw,inner sep=1.5pt] (char) {\footnotesize#1};}}

\renewcommand{\arraystretch}{1.2}

\definecolor{bg}{rgb}{0.95,0.95,0.95}
\definecolor{bestres}{HTML}{D6DEFC}

\title{Small Languages, Big Models:\\ A Study of Continual Training on Languages of Norway}

\author{David Samuel\hspace{1em}Vladislav Mikhailov\hspace{1em}Erik Velldal\hspace{1em}
Lilja Øvrelid\vspace{-0.4em}\\
\textbf{Lucas Charpentier\hspace{1em}Andrey Kutuzov\hspace{1em}Stephan Oepen}\vspace{0.2em}\\
University of Oslo, Language Technology Group\vspace{0.2em}\\
\texttt{davisamu@ifi.uio.no}
}

\date{}

\begin{document}
\maketitle

\begin{abstract}
Training large language models requires vast amounts of data, posing a challenge for less widely spoken languages like Norwegian and even more so for truly low-resource languages like Northern Sámi. To address this issue, we present a novel three-stage continual training approach that substantially improves the downstream performance together with the inference efficiency for the target languages. Based on our findings, we train, evaluate, and openly release a new generative language model for Norwegian Bokmål, Nynorsk, and Northern Sámi with 11.4 billion parameters: \textit{NorMistral-11B}.
\vspace{1.5em}
\end{abstract}

\section{Introduction}

The development of large language models typically requires massive amounts of training data, which benefits wide-spread languages such as English, but poses a significant challenge for less widely spoken languages. Norwegian, with its two written standards Bokmål and Nynorsk,\footnote{While Bokmål is the main variety, roughly 15\% of the Norwegian population uses Nynorsk. 
The two varieties are so closely related that they may be regarded as ‘written dialects’, but the lexical differences can be relatively large.} currently has approximately 24B words available in our filtered text collection -- about three orders of magnitude less than English \citep{penedo2024the}. The situation is even more challenging for Northern Sámi with only 40 million words available.\footnote{The Sámi languages are a group of Uralic languages, of which Northern Sámi is the most widely used variant. With the number of speakers estimated to be between 15,000 and 25,000 in total across Norway, Sweden and Finland, it is still considered to be an endangered language. As the Sámi people are recognized as an Indigenous people in Norway, Sámi has status as an official language along with Norwegian.}

To address this data scarcity, we propose a novel approach combining three key elements: knowledge transfer from existing models, data augmentation with related languages, and targeted upsampling. This method enables us to train an 11.4B-parameter model that achieves state-of-the-art performance across Norwegian language tasks while obtaining strong capabilities in Northern Sámi. The three main research contributions of this paper can be summarized as follows:
\begin{enumerate}
  \item \textbf{Novel training method for data-constrained language models}\hspace{1.5em}We propose a three-stage training method for efficient adaptation of existing language models to lower-resource languages. Our results demonstrate that this approach works well for adapting a Mistral model to Bokmål, Nynorsk and Northern Sámi. Our model achieves the \textit{state-of-the-art} performance on tasks requiring deep linguistic understanding and world knowledge in Norwegian contexts -- while being more than 30\% faster than the original Mistral model on Norwegian inputs.
  \item \textbf{Flexible masked-causal model}\hspace{1.5em}We train a general language model that can act as a causal generative model as well as a fully-bidirectional embedding model. This approach allows it to be used as any other generative model while allowing future usage as a finetuned encoder model.
  \item \textbf{Truly open source}\hspace{1.5em}We openly release NorMistral-11B under a permissive Apache 2.0 license -- {\footnotesize\textbf{\url{https://hf.co/norallm/normistral-11b-warm}}} -- as well as three smaller 7B-parameter models and a new corpus for Northern Sámi. The model is trained on fully transparent corpora and evaluated on a robust set of prompts that are included in the paper. The training and evaluation scripts are available at {\footnotesize\textbf{\url{https://github.com/ltgoslo/norallm}}}.
\end{enumerate}

\noindent

\begin{figure*}[th!]
    \centering
    \includegraphics[width=1.0\linewidth]{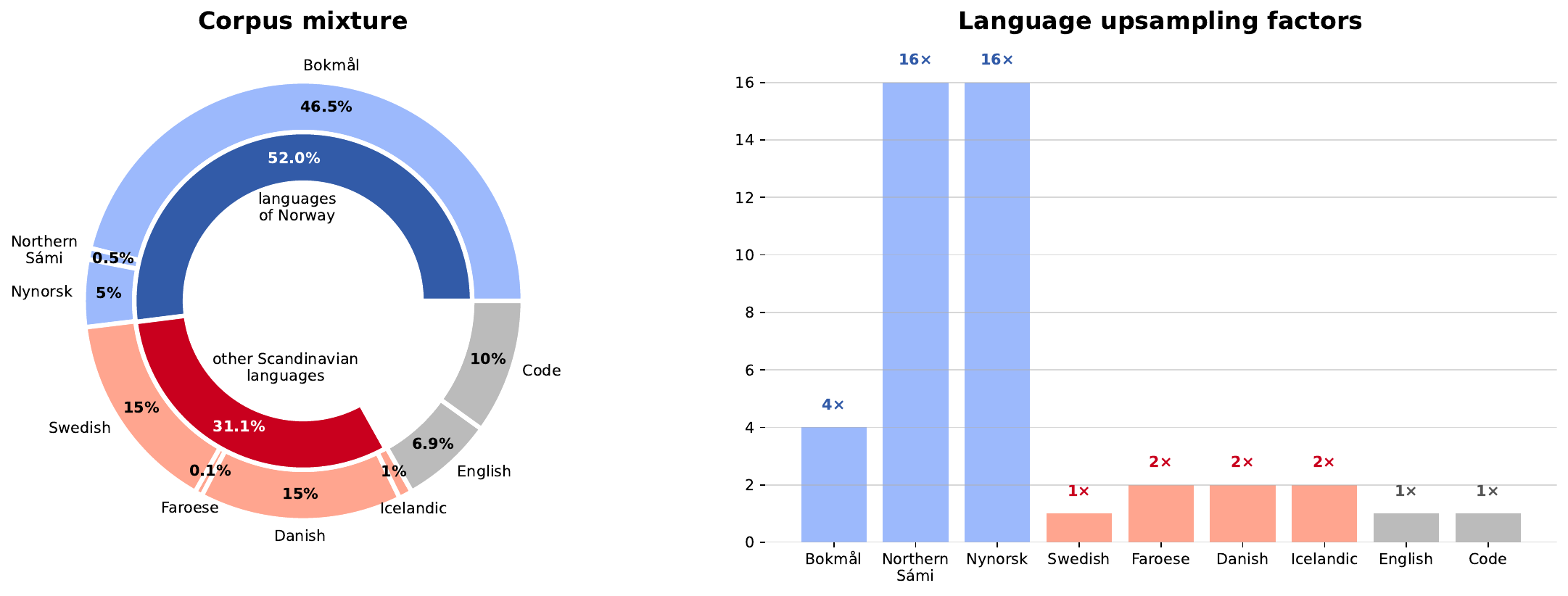}
    \caption{\textbf{Language composition of training corpus}\hspace{1em}The left figure shows the proportions of languages in the final corpus mixture, with the target languages of Norway in blue, related languages in red, and other data sources in gray. The right figure then displays the upsampling factors used to get the aforementioned proportions.}
    \label{fig:corpus}
\end{figure*}

\noindent
In the following sections, we first describe the training corpus of NorMistral-11B in \Cref{sec:corpus}; then the training and evaluation methodology of this model in \Cref{sec:method}. In \Cref{sec:results}, we then evaluate this model and compare it against other existing models. The following \Cref{sec:comparisons} then goes into more detail by testing the training choices in our methodology.\Cref{sec:related} describes previous works that inspired this paper. Additional appendices then offer further analyses and a detailed description of the evaluation setup.

\vspace{0.75em}
\section{Training corpus} 
\label{sec:corpus}

Our goal is to train a model for the official languages of Norway. However, this task is made difficult by the uneven distribution of these languages and the fact that there is only about 24 billion words in these languages available in the publicly accessible high-quality corpora (see below).

\vspace{0.25em}
\subsection{Combating the data constraints}

24B words is about three orders of magnitude less than what is currently available for English language models \citep{penedo2024the}. Assuming the Chinchilla scaling laws \citep{NEURIPS2022_c1e2faff}, we could `optimally' train only a 1-billion-parameter model on such a small dataset. However, we are able to train a much larger model due to: \circled{1} transferring knowledge from a model already trained on a large English-centric corpus; \circled{2} augmenting the corpus with other related Scandinavian languages (Danish, Swedish, Icelandic, and Faroese), as well as English and programming code \citep{Luukkonen2024Poro3A}; \circled{3} further increasing the size by repeating the data in target languages -- this follows the data-constrained scaling laws by \newcite{muennighoff2023scaling}, which showed that four repetitions do not have any noticeable negative effects on the regular scaling laws. The resulting corpus of 250B non-unique tokens is then `compute-optimal' for the 11.4B parameters of our model \citep{NEURIPS2022_c1e2faff}. In the previous work, the NorGPT-23B LLM trained on the available Norwegian data by \newcite{liu-etal-2024-nlebench} did not outperform smaller 3B models. Similarly, for Finnish, \newcite{luukkonen-etal-2023-fingpt} reported a decrease in performance when moving from 8B to 13B parameters on a similarly sized corpus. These observations support our decision not to move beyond the 11B size. 

\vspace{0.25em}
\subsection{Combating the uneven distribution} We target Norwegian and Sámi, the two official languages of Norway. Specifically, we target the \textit{Bokmål} written variant of Norwegian with 24 billion words in our corpus, the \textit{Nynorsk} variant with 0.5 billion words, and \textit{Northern Sámi}, which has only 40 million words in our corpus collection. To mitigate the large size differences, we further upsample the two lower-resource languages \citep{conneau-etal-2020-unsupervised}. To avoid overfitting on many repetitions of the same data, we follow the experimental results in \newcite{muennighoff2023scaling} and repeat the data at most 16 times. This approach yields the final language proportions shown in \Cref{fig:corpus}.

\begin{figure*}[ht!]
    \centering
    \includegraphics[width=\textwidth]{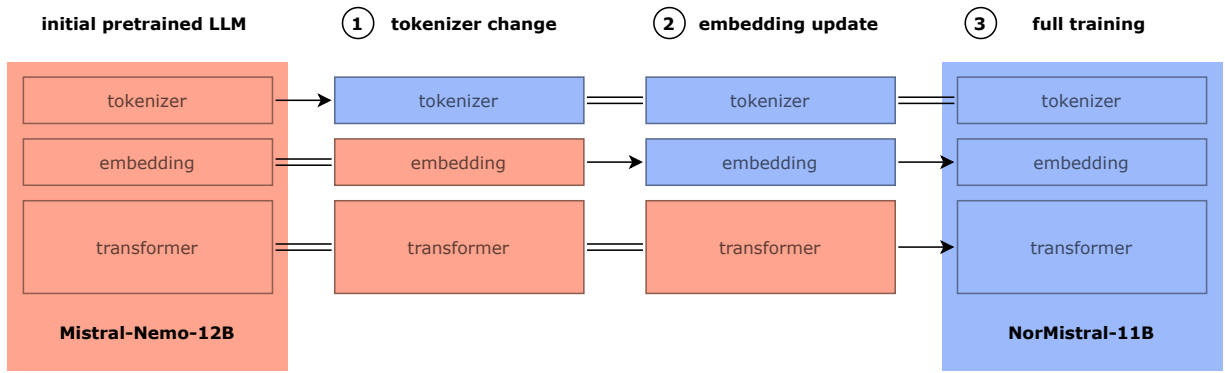}
    \caption{\textbf{Three-stage continual pretraining}\hspace{1em}We propose a novel continual pretraining pipeline consisting of \circled{1} creating a new tokenizer optimized for the training corpus, \circled{2} realigning the embedding weights to the new tokens, and \circled{3} training the full language model. Arrows symbolize changes between stages, while double-lines represent no changes.}
    \label{fig:training-pipeline}
\end{figure*}

\vspace{0.25em}
\subsection{Data sources} 

\paragraph{Existing corpora}
We source most of the data from existing publicly available corpora: \circled{1} Bokmål and Nynorsk filtered from the public sources with permissive licenses from the \textit{Mímir Core} corpus from \newcite{delarosa2025impact}, which itself consists of the Norwegian Colossal Corpus \citep[NCC;][]{kummervold-etal-2022-norwegian}, CulturaX \citep{nguyen2023culturax}, and the HPLT corpus v1.2 \citep{de-gibert-etal-2024-new}; \circled{2} Bokmål, Nynorsk, Swedish, Danish, and Icelandic from \textit{CulturaX} \citep{nguyen2023culturax}; \circled{3} high-quality English from \textit{FineWeb-edu} \citep{penedo2024the}; \circled{4} code from the high-quality part of \textit{Stack v2} \citep{lozhkov2024starcoder}; 
\circled{5} Faroese and Northern Sámi from \textit{Glot500} \citep{imanigooghari-etal-2023-glot500}; and \circled{6} Northern Sámi from the \textit{SIKOR free corpus} \citep{8AK7KZ_2016}. 

\paragraph{Web crawl for Sámi} The only exception to using existing resources is a part of the Sámi corpus. To obtain more texts for this low-resource language, we conducted a web crawl through admissible web pages in Northern Sámi. The crawl was seeded from the external links of the Sámi Wikipedia and continued with a breadth-first search through webpages that were identified as Northern Sámi using GlotLID \citep{kargaran-etal-2023-glotlid} and that allowed crawling according to their Robots Exclusion Protocol. The raw HTML documents were converted into natural text using Trafilatura \citep{barbaresi-2021-trafilatura}. We have published the web-crawled texts (fuzzy deduplicated at the document level) online at {\footnotesize\textbf{\url{https://hf.co/datasets/ltg/saami-web}}}. In total, it contains about 13 million whitespace-separated words.

\vspace{0.75em}
\section{Training and evaluation of NorMistral}
\label{sec:method}

This section describes the training and evaluation pipeline of NorMistral-11B; a continually trained \texttt{Mistral-Nemo-Base-2407} language model.\footnote{Available on HuggingFace at \url{hf.co/mistralai/Mistral-Nemo-Base-2407}} The presented methods are evaluated later in \Cref{sec:comparisons}.

\vspace{0.25em}
\subsection{Three-stage continual pretraining}
Our aim is to model three lower-resource languages. To achieve this, we rely on models initially trained on more resource-rich languages and continually train them on our corpus. In order to get a model that works efficiently for the target language,  
we propose a novel three-stage training process, which consists of tokenizer change, embedding update, and full training (\Cref{fig:training-pipeline}).

\paragraph{Stage 1: Tokenizer change}
Before training the language model, we create a new subword tokenizer optimized for the target distribution of languages. While keeping the original tokenizer might not necessarily worsen performance, the main goal of this step is to improve the efficiency of training and inference. As evident from  \Cref{tab:tokenizers}, the new tokenizer produces 30\% shorter sequences on average, which translates to more than 30\% faster inference time; while requiring 800 million less parameters due to the smaller vocabulary size. We measure the inference speed-up on a downstream task in \Cref{app:efficiency}, confirming the theoretical benefits.

The tokenizer is optimized for the entire training corpus via the greedy byte-pair encoding algorithm \citep[BPE;][]{Gage1994ANA}. We use the same tokenizer definition as the original Mistral-Nemo-12B: byte-level BPE tokenizer without any Unicode normalization and with a fairly complex pretokenizer regular expression. The pretokenizer splits numbers into individual digits as in \newcite{10.5555/3648699.3648939}. Note that the tokenization is completely lossless and reversible as out-of-vocabulary characters can be split into individual UTF-8 bytes that are always in-vocabulary as atomic tokens.

\paragraph{Stage 2: Embedding update}
Since all tokens are changed in the previous stage, we need to update the input and output embedding weights next. While it is possible to skip this stage and simply continue training the full model, misaligned embeddings lead to a large initial loss spike, to large (essentially random) gradients for the non-embedding parameters, and thus to catastrophic forgetting \citep{MCCLOSKEY1989109}. Instead, we follow the tokenizer adaptation method by \newcite{de-vries-nissim-2021-good}, aligning the embedding parameters by continually training the language model for 1\,000 steps with frozen non-embedding parameters.

\begin{figure*}[ht!]
    \centering
    \includegraphics[width=\textwidth]{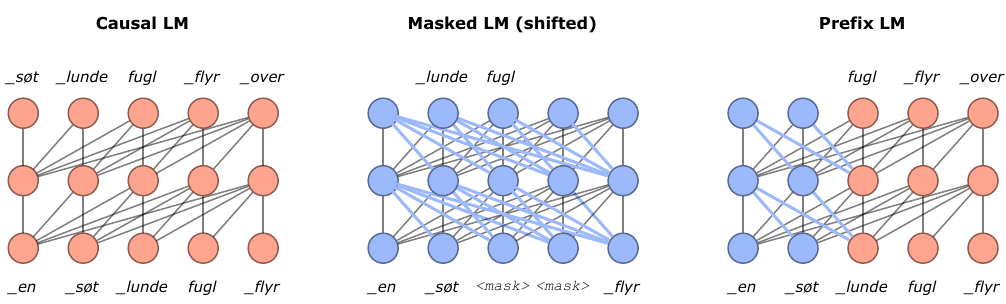}
    \caption{\textbf{Inference modes of NorMistral-11B}\hspace{1em}The hybrid masked-causal pretraining allows the model to be more flexible during inference. It can not only serve as a unidirectional causal language model (left), but also as a fully bidirectional masked language model (middle), or as a partially bidirectional prefix language model (right). The diagrams illustrate possible attention connections.}
    \label{fig:hybrid}
\end{figure*}

The initial token embeddings are transferred from the original embedding matrix \citep{gu-etal-2018-universal, wang-etal-2019-improving}. Since we use the same tokenizer type as the original Mistral model, many tokens are present in both vocabularies; the embeddings for these are initialized by as direct copies of the original vectors. Tokens not present in the original vocabulary are tokenized (with the original tokenizer) to obtain sub-tokens within the vocabulary; the embedding vectors are then initialized by taking the average of the sub-token embeddings. 

\paragraph{Stage 3: Full training} After realigning the embedding vectors, we continue by unfreezing the remaining parameters and training the full model.

The transformer architecture is inherited from the original Mistral model \citep{Jiang2023Mistral7}, which is based on the improved Llama architecture \citep{Touvron2023LLaMAOA}. This mainly entails: \circled{1} pre-normalization with the RMSNorm function for improved training stability \citep{nguyen-salazar-2019-transformers, zhang-sennrich-neurips19}, \circled{2} SwiGLU activation function for improved expressive power of the feed-forward modules \citep{Shazeer2020GLUVI}, \circled{3} rotary positional embeddings for their ability to generalize to longer sequences \citep{su2023roformerenhancedtransformerrotary, liu2024scaling}, and \circled{4} grouped-query attention for improved inference efficiency \citep{ainslie2023gqa}. The remaining architectural details are based on the original transformer design by \newcite{NIPS2017_3f5ee243}. The hidden dimension is set to 5\,120, the intermediate one to 14\,336, and there are 40 layers in total. The attention modules have 32 query heads and 8 key \& value heads, each of dimension 128. There are 51\,200 tokens in the subword vocabulary. 

\begin{table}[t!]
\resizebox{\columnwidth}{!}{%
\begin{tabular}{@{}l@{\hspace{3em}}r@{\hspace{2em}}ccc@{}}
\toprule
\textbf{Tokenizer} & \textbf{\# tokens} & \textbf{\textsc{nob}} & \textbf{\textsc{nno}} & \textbf{\textsc{sme}} 
\\ \midrule
Mistral-Nemo-12B  & 131\,072  & 1.79 & 1.87  & 2.63 
\\ 
NorMistral-11B & 51\,200 & 1.22 & 1.28 & 1.82 
\\
\bottomrule
\end{tabular}%
}
\caption{\textbf{Tokenizer statistics}\hspace{1em}The vocabulary size and subword-to-word split ratios of different tokenizers for Bokmål (\textsc{nob}), Nynorsk (\textsc{nno}) and Northern Sámi (\textsc{sme}). Lower split ratios result in shorter subword sequences and thus in faster training and inference.}
\label{tab:tokenizers}
\end{table}

\renewcommand{\arraystretch}{1.25}

\begin{table*}[!th]
\resizebox{\textwidth}{!}{%
\begin{tabular}{@{}l@{\hspace{0em}}c@{\hspace{3em}}
>{\columncolor[HTML]{D6DEFC}}c cccccccc@{}}
\toprule
  \textbf{Benchmark} &
  \textbf{Language} &
  \rotatebox{90}{\textbf{NorMistral-11B}} &
  \rotatebox{90}{\textbf{NorwAI-Mistral-7B}} &
  \rotatebox{90}{\textbf{NorwAI-Llama2-7B}} &
  \rotatebox{90}{\textbf{NorMistral-7b-warm}} &
  \rotatebox{90}{\textbf{NorGPT-3B}} &
  \rotatebox{90}{\textbf{Viking-7B}} &
  \rotatebox{90}{\textbf{Viking-13B}} &
  \rotatebox{90}{\textbf{Mistral-Nemo-12B}} \\ \midrule
\textsc{\textsubscript{Reading comprehension}} & & \\
\hspace{1em}Belebele \textsubscript{(0-shot)}      & Bokmål              & 56.7          & 33.4   & 38.0 & 37.4 & 26.8       & 27.6 & 28.2          & \textbf{62.8} \\
\hspace{1em}NorQuAD \textsubscript{(1-shot)}          & Bokmål              & \textbf{76.7} & 63.0   & 39.2 & 64.8 & 3.0   & 48.4 & 57.1          & 76.5          \\[0.5em]
\textsc{\textsubscript{Sentiment analysis}} & &  \\
\hspace{1em}NoReC \textsubscript{(sentence-level; 16-shot)}           & Bokmål              & \textbf{90.5} & 88.6 & 86.0 & 84.9 & 49.7         & 77.9 & 79.2          & 86.9          \\
\hspace{1em}NoReC \textsubscript{(document-level; 1-shot)} & Bokmål & \textbf{91.2} & 81.2 & 79.2 & 82.9 & 51.5 & 80.4 & 86.8 & 89.2 \\[0.5em]
\textsc{\textsubscript{Commonsense reasoning}} & & \\
\hspace{1em}NorCommonsenseQA \textsubscript{(0-shot)} & Bokmål              & \textbf{61.0} & 54.2  & 49.7 & 51.3 & 34.7     & 44.9 & 51.1          & 46.9          \\
\hspace{1em}NorCommonsenseQA \textsubscript{(0-shot)} & Nynorsk             & \textbf{51.6} & 43.2  & 37.9 & 43.2 & 29.5        & 39.0 & 40.0          & 33.7          \\[0.5em]
\textsc{\textsubscript{World knowledge}} & & \\
\hspace{1em}NRK-Quiz-QA \textsubscript{(0-shot)}         & Bokmål              & \textbf{63.7} & 55.2  & 52.3 & 57.9 & 33.1      & 44.2 & 51.0          & 47.4          \\
\hspace{1em}NRK-Quiz-QA \textsubscript{(0-shot)}         & Nynorsk             & \textbf{71.9} & 65.2  & 64.3 & 65.9 & 37.3       & 51.1 & 54.8          & 47.2          \\
\hspace{1em}NorOpenBookQA \textsubscript{(16-shot)}   & Bokmål              & 77.9 & 52.3 & 52.3 & 49.0 & 29.5 & 48.7 & 47.0 & \textbf{86.9}      \\
\hspace{1em}NorOpenBookQA \textsubscript{(16-shot)}   & Nynorsk             & 77.8 & 45.6   & 38.9 & 41.1 & 34.4 & 27.8 & 36.7 & \textbf{86.7}       \\[0.5em]
\textsc{\textsubscript{Summarization}} & & \\
\hspace{1em}NorSumm \textsubscript{(0-shot)}          & Bokmål              & \textbf{45.0} & 12.2  & 10.7 & 16.5 & 33.8       & 31.9 & 36.3         & 44.9          \\
\hspace{1em}NorSumm \textsubscript{(0-shot)}          & Nynorsk             & \textbf{32.6} & 10.3 & 10.4 & 8.6 & 24.3       & 25.7 & 28.8          & 30.9          \\[0.5em]
\textsc{\textsubscript{Grammatical error correction}} & &  \\
\hspace{1em}ASK-GEC \textsubscript{(16-shot)}         & Bokmål              & 52.6          & \textbf{53.2} & 51.4 & 48.7 & 1.8 & 51.1 & 52.4          & 43.9          \\[0.5em]
\textsc{\textsubscript{Language identification}} & &  \\
\hspace{1em}SLIDE \textsubscript{(16-shot)}         & $\substack{\text{Bokmål, Nynorsk,} \\ \text{Danish, Swedish}}$              & \textbf{98.2}          & 95.7 & 93.5 & 98.1 & 40.3 & 77.2 & 84.4 & 87.3     \\[0.5em]
\textsc{\textsubscript{Translation}} & & \\
\hspace{1em}Tatoeba \textsubscript{(from English; 16-shot)}         & Bokmål  & 58.8 & 58.7 & 57.9 & 57.2 & 1.8 & 59.7 & \textbf{60.0} & 49.6        \\
\hspace{1em}Tatoeba \textsubscript{(from English; 16-shot)}         & Nynorsk &  \textbf{48.0} & 47.4 & 47.4 & 44.7 & 2.6 & 45.6 & 45.6 & 35.7     \\
\hspace{1em}Tatoeba \textsubscript{(from English; 16-shot)}         & Northern Sámi    &       \textbf{50.4} & 27.5 & 28.5 & 18.5 & 0.0 & 7.8 &       11.6          &    6.5       \\ \bottomrule
\end{tabular} %
}
\caption{\textbf{Performance of NorMistral-11B}\hspace{1em} This table compares the performance of NorMistral-11B to the performance of other dense generative models that support Norwegian. All models are evaluated with the same fully-causal in-context-learning setup without any parameter updates. The best results are in bold; higher values are always better. The performance is evaluated by accuracy (Belebele, NorCommonsenseQA, NorOpenbookQA \& NRK-Quiz-QA), F\textsubscript{1} score (NorQuAD \& NoReC), ROUGE-L (\citealp{lin-2004-rouge}; NorSumm), ERRANT F\textsubscript{0.5} (\citealp{bryant-etal-2017-automatic}; ASK-GEC), accuracy (SLIDE), and BLEU (\citealp{papineni-etal-2002-bleu}; Tatoeba). We report the maximum performance score across all prompts. The random guessing baselines are 20\% for NorCommonSenseQA, 25\% for Belebele and NorOpenBookQA, 28\% / 27\% for NRK-Quiz-QA \textsc{nob} / \textsc{nno}, and 48.5\% / 48.4\% for NoReC sentence-level / document-level.
}
\label{tab:results}
\end{table*}

We trained the model on 250 billion tokens, which equates to 60\,000 steps of 1\,024 $\times$ 4\,096 tokens (number of samples $\times$ sequence length). We used the trapezoidal learning-rate schedule with a peak learning rate of $1\cdot10^{-4}$, 1\,000 warm-up steps and 10\,000 decay steps; this schedule allows for further pretraining of this model on more tokens in the future \citep{hagele2024scaling}. The optimization was performed using AdamW \citep{loshchilov2018decoupled}, with $\beta_1=0.9$, $\beta_2=0.95$, $\epsilon=10^{-8}$, and weight decay of $0.1$. No dropout was applied.

The computations were conducted on 256 AMD MI250X GPUs and used 55\,000 GPU hours in total -- which equals to 8.5 days of runtime on the distributed setup. The model was trained with a reduced \texttt{bfloat16} precision and the parameters were sharded with model parallelism -- pipeline parallelism of 2, tensor parallelism of 2, and a zero-redundancy optimizer \citep{shoeybi2020megatronlmtrainingmultibillionparameter, 10.1145/3394486.3406703, 10.5555/3433701.3433727}. The overall theoretical computation cost of the training was $1.7\cdot10^{22}$ FLOPs, with an average of 38\% model FLOP/s utilization (MFU) on the actual hardware.

\vspace{0.25em}
\subsection{Hybrid masked-causal language modeling}
\label{sec:hybrid}

While causal LMs have recently become very popular, the limited unidirectional text processing limits their learning abilities \citep{DBLP:journals/corr/abs-2311-07468} and expressive power \citep{ewer2024entpencoderonlytokenprediction}; especially for finetuning \citep{devlin-etal-2019-bert, 10.5555/3455716.3455856}. Furthermore, it has been recently demonstrated that fully-bidirectional masked models share the same generative abilities, but without limitations of causal models \citep{samuel2024berts}. Following this observation, we train a model that can be flexibly used as a masked or causal language model.

\paragraph{Training objective}
We combine two training objectives during pretraining
: the standard causal language modeling one as well as masked next-token prediction \citep[MNTP;][]{DBLP:journals/corr/abs-2311-07468, behnamghader2024llmvec}, a variation of masked language modeling where the next token is predicted rather than the current one (see Masked LM (shifted) in \cref{fig:hybrid}). This has been used by \citet{gptbert}, with evidence of providing better causal modeling quality and increased finetuning performance. We trained with 90\% causal LM and 10\% MNTP. This ratio is rather conservative -- to teach the model bidirectional processing without drifting too much from its original training objective.

\vspace{0.25em}
\subsection{Experimental Setup}
\label{sec:setup}
We compare the performance of NorMistral-11B with publicly available LMs using \texttt{NorEval},\footnote{\href{https://github.com/ltgoslo/noreval}{\texttt{github.com/ltgoslo/noreval}}} an open-source framework for evaluating Norwegian generative LMs built on \texttt{lm-evaluation-harness} \cite{eval-harness}. The evaluation is run in $k$-shot scenarios with $k \in \{0, 1, 16\}$ on ten benchmarks. We report the maximum $k$ for each benchmark across a set of prompts, which depends on the availability of a training/development set for demonstration examples and on the average length of these examples. 

\paragraph{Baselines} We use seven pretrained LMs of comparable size accessed via the \texttt{Transformers} library \cite{wolf-etal-2020-transformers} as our baselines: \texttt{Norw\-AI\--Mistral\--7B}, \texttt{Norw\-AI\--Llama\-2\--7B}, \texttt{nor\-mistral\--7\-b\--warm}, \texttt{Nor\-GPT\--3\-B} \citep{liu-etal-2024-nlebench}, \texttt{Viking\--7B}, \texttt{Viking\--13\-B}, and \texttt{Mistral\--Nemo\--12\-B}.

\paragraph{Benchmarks} The models are evaluated only on datasets created by native speakers. We consider the following language understanding and generation tasks: \circled{1} reading comprehension (NorQuAD; \citealp{ivanova-etal-2023-norquad} \& Belebele; \citealp{bandarkar-etal-2024-belebele}), \circled{2} sentiment analysis (NoReC; \citealp{velldal-etal-2018-norec}), \circled{3} commonsense reasoning (NorCommonsenseQA; \citealp{mikhailov2025collection}), \circled{4} world knowledge (NRK-Quiz-QA \& NorOpenBookQA; \citealp{mikhailov2025collection}), \circled{5} summarization (NorSumm; \citealp{touileb2025benchmarking}), \circled{6} grammatical error correction (ASK-GEC; \citealp{jentoft2023grammatical}), \circled{7} language identification (SLIDE; \textbf{\footnotesize\url{https://github.com/ltgoslo/slide}}), and \circled{8} translation (Tatoeba; \citealp{tiedemann-2020-tatoeba}). \texttt{NorEval} provides a set of task-specific 4--6 prompts written by Norwegian native speakers, which allows to account for prompt sensitivity \cite{lu-etal-2024-prompts}. More details about each task with a complete list of prompts are given in \Cref{appendix:evaluation}.

\vspace{0.75em}
\section{Results}
\label{sec:results} 

We report the aggregated evaluation results in \Cref{tab:results} and fine-grained evaluation results in \autoref{appendix:evaluation}. Overall, we see a positive indication of NorMistral-11B being a strong Norwegian model as it outperforms other evaluated systems on the majority of tasks.


\paragraph{Comparison to the base model} Even though Mistral-Nemo-12B is an English-centric model, it performs well on the Norwegian benchmarks even before any continual pretraining. While we see a clear increase in performance after further training when evaluated on native Norwegian datasets, there is a notable decrease in performance on Belebele (a well-known multilingual dataset) and NorOpenBookQ (an adaptation of a popular English benchmark). This aspect requires a further study, but overall, we believe that the results clearly show the benefit of three-stage continual pretraining.

\paragraph{Bokmål, Nynorsk and Sámi performance} We evaluate the models on all target languages: Bokmål, Nynorsk and Northern Sámi. Relative to other models, the performance gains of NorMistral-11B stay consistent across these three languages.

It is possible to estimate the difference in performance on Nynorsk compared to Bokmål when focusing on NorSumm, a dataset that is perfectly balanced and parallel for the two variants of Norwegian. The substantially higher score for Bokmål indicates that the much smaller amount of Nynorsk in the training corpus (even after upsampling) limits the downstream performance on this language variant.

The results on the English-to-Sámi translation suggest that our model was able to learn aspects of this language even though it made only 0.5\% of the training corpus. However, any stronger claim about the level of understanding of Sámi would require a substantially more robust benchmarking suite than what is currently available.

\vspace{0.25em}
\subsection{Using NorMistral in practice} Large language models can be utilized in many different ways. We used the most direct and straightforward one for comparing Norwegian models -- in-context learning -- but there is a broader spectrum of methods with varying complexity-to-performance trade-offs. We evaluate the most common methods in \Cref{tab:norquad} using NorQuAD:

\paragraph{In-context learning}
This is the most popular method of using large language models, mostly because it does not require any further training \citep{NEURIPS2020_1457c0d6}. Using just one sample from the training set as a demonstration can substantially improve the output quality on NorQuAD. More demonstrations can improve the performance further, but at the cost of reduced inference speed.

\paragraph{Quantization}
In order to reduce the large memory cost of large language models, a popular method is reducing the precision of their parameters. Specifically, we test 8-bit and 4-bit quantization \citep{dettmers2022gptint, pmlr-v202-dettmers23a}. There is no noticeable decrease of performance on NorQuAD when lowering the precision from the original 16 bits. Note that some GPUs can also increase their throughput at the lowered precision.

\paragraph{Full finetuning} The best-performing strategy is to do supervised finetuning of all learnable parameters. This method is also the most difficult to set up, the large memory requirements necessitate distributed training with some model sharding. However, after finetuning, this method clearly outperforms all other ones without any additional cost. Interestingly, when finetuned with partially-bidirectional attention masks (as a prefix LM), the model even exceeds the estimated human performance on NorQuAD -- $91.1$ F\textsubscript{1} score and $78.1$ EM accuracy \citep{ivanova-etal-2023-norquad}.

\begin{table}[t]
\resizebox{\columnwidth}{!}{%
\begin{tabular}{@{}l@{\hspace{2em}}ccc@{}}
\toprule
\textbf{Method} &
  \textbf{F\textsubscript{1}} &
  \textbf{EM} &
  \textbf{\begin{tabular}[c]{@{}c@{}}Runtime\\[-0.5em]train / eval\end{tabular}} \\ \midrule
0-shot (causal)   &        59.7              &            33.5         & \phantom{0}\textbf{0} / \phantom{0}\textbf{6} min            \\
1-shot (causal)   &     76.7                &          55.3            & \phantom{0}\textbf{0} / \phantom{0}8 min           \\
8-shot (causal) & 79.6                 & 60.8                 & \phantom{0}\textbf{0} / 23 min           \\[0.5em]
0-shot (4-bit, causal) &    59.2              &            33.5         & \phantom{0}\textbf{0} / \phantom{0}\textbf{6} min \\
0-shot (8-bit, causal) & 59.1 & 33.7 & \phantom{0}\textbf{0} / \phantom{0}\textbf{6} min \\[0.5em]
Full finetuning (causal) &  90.4 & 79.2 & 57 / \phantom{0}\textbf{6} min \\
Full finetuning (prefix)  &  \textbf{92.2} & \textbf{80.3} & 57 / \phantom{0}\textbf{6} min \\ [0.5em]
LoRA finetuning (causal) & 89.9 & 77.1 & 18 / \phantom{0}\textbf{6} min \\
LoRA finetuning (prefix) & 91.3 & 79.0 & 18 / \phantom{0}\textbf{6} min \\ \bottomrule
\end{tabular} %
}
\caption{\textbf{Evaluation methods}\hspace{1em}NorMistral-11B can be flexibly used in many different ways for solving downstream tasks. We compare them on NorQuAD, a dataset for extractive question answering. NorMistral can be finetuned as a standard causal language model and also as a partially bidirectional prefix language model. We also show the total training and evaluation time for each method (run on AMD MI250X GPUs). We use the two standard metrics for extractive question answering: F\textsubscript{1} score and exact-match accuracy (EM).}
\label{tab:norquad}
\end{table}

\paragraph{LoRA finetuning}
Further training NorMistral on a downstream task is more demanding, but it is the preferred way for achieving the best performance -- as long as there is a sizeable training set available. Low-rank adaptation (LoRA) reduces the computational cost of finetuning by freezing all original model parameters and training only small low-rank adaptors \citep{hu2022lora}. The resulting model is 10 F\textsubscript{1} percentage points better than the best few-shot prompt while running almost 4 times faster because of shorter context lengths. Because of its hybrid pretraining (\Cref{sec:hybrid}), NorMistral can also be finetuned as a partially-bidirectional prefix language model, which further improves its performance by $1.4$ points without any additional computational cost.

\vspace{0.75em}
\section{Methodological comparisons}
\label{sec:comparisons}

We have conducted an initial comparative study of different training methods before settling on the pretraining process from \Cref{sec:method} and training NorMistral-11B. The results are presented in \Cref{tab:comparisons}, where different models are evaluated on a representative subset of available Norwegian benchmarks: extractive question answering (1-shot NorQuAD), binary sentence-level polarity classification (16-shot NoReC), world knowledge (0-shot NRK-Quiz-QA) and machine translation (16-shot English-to-Bokmål Tatoeba).

\paragraph{Architectural choice} There are many promising improvements of the original GPT neural architecture \citep{Radford2018ImprovingLU} -- we considered two recent and well-studied architectures: BLOOM \citep{bigscience_workshop_2022} and Llama \citep{Touvron2023LLaMAOA}, which is also used for training the Mistral models \citep{Jiang2023Mistral7}. We adopted the training hyperparameters suggested by the respective papers and trained two models with 7 billion parameters on the same Norwegian corpus and with the same Norwegian tokenizer. \Cref{tab:comparisons} clearly shows that the Llama architecture is preferred for our training corpus and Norwegian benchmarks.

\paragraph{From scratch vs. warm-starting} The central research question of this paper is how to train a good large language model for relatively small languages. Here we test our proposed three-stage continual pretraining and compare it against a model trained from scratch. For a fair comparison, we train two 7-billion-parameter models on the same Norwegian corpus (the Norwegian Colossal Corpus by \citealp{kummervold-etal-2021-operationalizing}), and with the same architecture and tokenizer. Note that we do not consider existing methods that do not adapt the subword vocabulary -- like simple continual training or adapter tuning \citep{yong-etal-2023-bloom} -- because they necessarily lead to inefficient inference (\Cref{tab:tokenizers}). The results in \Cref{tab:comparisons} demonstrate that the knowledge transfer from an English-centric model works and the model is able to be adapted to new languages.

\paragraph{Hybrid masked-causal modeling} 

Interestingly, we do not observe an overall increase in performance after training with the `dual' training objective, as opposed to the observations by \citet{gptbert}. However, we believe that this can be explained by continued training -- the hybrid masked-causal training is used for a negligable number of steps compared to the fully-causal pretraining of the base Mistral model.

\begin{table}[t!]
\centering
\resizebox{\columnwidth}{!}{%
\begin{tabular}{@{}l@{\hspace{0.5em}}c@{\hspace{0.5em}}ccc@{}}
\toprule
\textbf{Training method} & \textbf{\begin{tabular}[c]{@{}c@{}}NorQuAD\\[-0.75em]\textsubscript{1-shot}\end{tabular}} & \textbf{\begin{tabular}[c]{@{}c@{}}NoReC\\[-0.75em]\textsubscript{16-shot}\end{tabular}} & \textbf{\begin{tabular}[c]{@{}c@{}}NRK\\[-0.75em]\textsubscript{0-shot}\end{tabular}} & \textbf{\begin{tabular}[c]{@{}c@{}}Tatoeba\\[-0.75em]\textsubscript{16-shot}\end{tabular}} \\ \midrule
\textsubscript{\textsc{transformer architecture}}  &  &  &  &  \\
\hspace{1em}BLOOM                            &  43.6 & 67.6 & 44.6 & 52.2 \\
\hspace{1em}Llama / Mistral                        & \textbf{43.7} & \textbf{80.3} & \textbf{48.2} & \textbf{53.4} \\[0.5em]
\textsubscript{\textsc{continual training}}        &  &  &  &  \\
\hspace{1em}init. from scratch            &  43.7 &  80.3 & 48.2 & 53.4  \\
\hspace{1em}three-stage continual         & \textbf{64.8}  & \textbf{84.9} & \textbf{57.9} &  \textbf{57.2} \\[0.5em]
\textsubscript{\textsc{hybrid training objective}} &  &  &  &  \\
\hspace{1em}causal-only                            & 67.0  & 86.0 & \textbf{59.0} & \textbf{58.8} \\
\hspace{1em}hybrid masked-causal                   & \textbf{69.3}  & \textbf{87.5} & 55.4  &  58.2 \\[0.5em]
\textsubscript{\textsc{training steps}}            &  &  &  &  \\
\hspace{1em}0 steps \textsubscript{(base model)} & 76.5 & 86.9 & 47.4 & 49.6 \\
\hspace{1em}0 steps \textsubscript{(adapted tokenizer)}                                &  73.5 & 89.4  & 44.2 & 51.4 \\

\hspace{1em}10,000 steps                                &  69.3 & 87.5  & 55.4  & 58.2 \\
\hspace{1em}20,000 steps                                & 70.5 & 89.2 & 57.7  & 58.8  \\
\hspace{1em}30,000 steps                                &  66.2 & 82.3 &  59.0 & 58.5 \\
\hspace{1em}40,000 steps                                &  68.5 &  87.0 & 61.1 & \textbf{58.9} \\
\hspace{1em}50,000 steps                               &  70.4 & 88.7 & 60.2 & 58.7  \\
\hspace{1em}60,000 steps                                & \textbf{76.7} & \textbf{90.5} & \textbf{63.7} & 58.8  \\ \bottomrule
\end{tabular}%
}
\caption{\textbf{Comparison of training methods}\hspace{1em}The methods are compared on NorQuAD with F\textsubscript{1} score, sentence-level Bokmål NoReC with F\textsubscript{1} score, Bokmål NRK-Quiz-QA with accuracy, and on English-to-Bokmål Tatoeba with BLEU.}
\label{tab:comparisons}
\end{table}

\paragraph{Number of training steps}

Finally, we compare the performance of model checkpoints saved at different points of training. We can make several observations from the results: \circled{1} they confirm the data-scaling laws by \citet{muennighoff2023scaling} as the model continues to improve even after (at least) four repetitions of the Norwegian data; \circled{2} tokenizer adaptation (the first two stages of our training method) is a simple and efficient way of adapting a model to a new language without losing performance; \circled{3} the three-stage continual pretraining does not affect all downstream tasks equally -- while it usually leads to monotonical improvement, there are some tasks (NorQuAD) that experience an initial decrease in performance. Further investigation is needed to determine if this drop is significant and if it can be avoided by a more careful switch to a new language distribution at the start of training.


\vspace{0.75em}
\section{Related work}
\label{sec:related}

 
\paragraph{Norwegian language models} 
There have been several prior efforts on creating language models for Norwegian. 
When it comes to creating openly available generative decoder-only models for Norwegian, most of the main efforts are listed in \Cref{sec:setup} and used in our experiments. However, one other notable mention is NB-GPT-J-6B -- a fine-tuned version of the English GPT-J-6B model.\footnote{\url{https://huggingface.co/NbAiLab/nb-gpt-j-6B}} Released by the National Library of Norway in 2022, it was the first large generative language model trained for Norwegian. 

There have also been several efforts on developing smaller transformer models, e.g., based on the BERT encoder architecture \citep{devlin-etal-2019-bert} and the T5 encoder-decoder architecture \citep{10.5555/3455716.3455856}. 
The NorBERT family of models were first released by \citet{kutuzov-etal-2021-large} and have by now reached their third iteration of releases \cite{samuel-etal-2023-norbench} and come in several different sizes; ranging fron 15M parameters for the XS model to 323M for NorBERT3 Large. \citet{samuel-etal-2023-norbench} also introduced the NorT5 family of models, ranging 32M to 808M parameters. 
Whereas the above-mentioned models where all trained from scratch for Norwegian, \citet{kummervold-etal-2021-operationalizing} trained NB-BERT (base and large) by fine-tuning the pre-trained mBERT model on Norwegian data, also reusing the tokenizer. A similar approach was followed for the North-T5 models.\footnote{\url{https://huggingface.co/north}} 

\paragraph{Language models for Northern Sámi} As for Northern Sámi, \newcite{Paul2024TowardsAM} has recently experimented with targeting this language. However, their models have not been published nor did they evaluate them on any downstream tasks; we are thus not able to compare them to our model. 

\paragraph{Continual training techniques} Adaptation of pretrained language models to new domains by continual training has a long history \citep{gururangan-etal-2020-dont}. Our three-stage continual pretraining is designed specifically for adapting language models to a new language -- by entirely replacing the original tokenizer, we can get an efficient model (by compressing the textual input into a short sequence of tokens) without the need of any extra parameters. Simple continual pretraining works well performance-wise but the training and inference computation cost is high \citep{ibrahim2024simple}. A substantially more efficient approach is to introduce a new tokenizer and replace the embedding layers (first two stages of our approach), as proposed by \citep{de-vries-nissim-2021-good, marchisio-etal-2023-mini}. Similarly, \newcite{Csaki2023EfficientlyAP} only use the first and last stage of our method -- they extend the vocabulary by 5\,000 new tokens and then train the full model. On the other hand, \newcite{kim2024efficienteffectivevocabularyexpansion} pursue a more careful approach, the most similar to our training method. They first extend the subword vocabulary with extra tokens and then meticulously train the new and old parameters in eight subsequent stages.

\vspace{0.75em}
\section{Conclusion}

We presented NorMistral-11B, a new large language model for Norwegian Bokmål, Nynorsk, and Northern Sámi. We proposed a novel three-stage continual pretraining approach that efficiently adapts existing models to other languages while maintaining high performance and increasing their inference speed. This approach involves training a new tokenizer, realigning embedding weights, and then training the full model. We also demonstrated the benefits of hybrid masked-causal pretraining, which allows the model to be used flexibly as either a causal or bidirectional model. Our extensive evaluation shows that NorMistral-11B achieves the state-of-the-art performance across a wide range of Norwegian  tasks, while also showing promising results for Northern Sámi. This suggests that our approach could be beneficial for developing large language models for other smaller languages. To facilitate further research and development, we have released NorMistral-11B, the three 7B models trained for \Cref{sec:comparisons}, training code, and a new Northern Sámi corpus at {\footnotesize\textbf{\url{https://github.com/ltgoslo/norallm}}}.



\vspace{0.75em}
\section*{Limitations} 

\paragraph{Limitations of the base language model}
Since NorMistral-11B is continually pretrained on the existing Mistral-Nemo-12B weights, the model is partially dependent on the training data of the original Mistral model. The exact composition of this training data is not known, which to some extent limits more detailed studies of this model. Specifically, the original model might have been trained on contaminated data, which could explain its high-scores on well-known evaluation tasks such as Belebele.

\paragraph{Computational cost}
As mentioned in \Cref{sec:method}, training NorMistral-11B took more than 55\,000 GPU/hours. This is a significant amount. We have not yet estimated the CO\textsubscript{2} footprint of the full training, but it was conducted on the LUMI supercomputer which is powered exclusively with renewable electricity and deployed in one of the most eco-efficient data centers in the world.\footnote{\url{https://www.lumi-supercomputer.eu/sustainable-future/}}

\paragraph{Evaluation of Northern Sámi knowledge}
Finally, our evaluation for Northern Sámi is limited to English-Sámi translation, which is obviously insufficient. Unfortunately, we lack more advanced or diverse benchmarks for low-resource languages like this one. We hope to see further development in this direction by the NLP community.


\vspace{0.75em}
\section*{Acknowledgments}

The computations were performed on resources provided through Sigma2 – the national research infrastructure provider for high-performance computing and large-scale data storage in Norway. We acknowledge Norway and Sigma2 for awarding this project access to the LUMI supercomputer, owned by the EuroHPC Joint Undertaking, hosted by CSC (Finland) and the LUMI consortium through project 465000498.

The efforts described in this paper were jointly funded by the University of Oslo and the HPLT project (High Performance Language Technologies; coordinated by Charles University).

The Norwegian part of our training corpus -- Mímir-core -- has been cleaned and graciously provided to us ahead of time by the National Library of Norway.


\bibliographystyle{acl_natbib}
\bibliography{nodalida2025, anthology_0, anthology_1}

\clearpage
\onecolumn
\appendix

\section{Inference efficiency of three-stage continual pretraining}
\label{app:efficiency}

In order to provide evidence for our claim that three-stage continual pretraining is necessary to increase the inference efficiency, we measure the actual inference speed on downstream tasks. Note that we specifically focus on the first stage of our pretraining recipe -- creating a brand new tokenizer for the target domain. Since in-context-lerning evaluation can be done in two modes -- classification or generation -- we measure the inference speed on both of them. Since the model quality might influence the number of generated tokens, we constrain the generation to only output tokens from the gold answers.

\paragraph{Speedup due to a new tokenizer} We compare the speed of the original language model, Mistral-Nemo-12B, with the speed of our model that was initialized from it, NorMistral 11B. The results in \Cref{tab:tokenizers-speed} show that completely changing the tokenizer results in a noticeable speed up in both tests.

\paragraph{Other evaluated models} For completeness, the inference speed of other models used in this paper are included as well; even though they have different number of non-embedding parameters or even completely different architectures. These additional measurements also show the benefit of replacing the entire vocabulary instead of only extending it with additional tokens.

\renewcommand{\arraystretch}{1.35}

\begin{table}[h]
\resizebox{\textwidth}{!}{%
\begin{tabular}{@{}lcl@{\hspace{1em}}ccc@{}}
\multicolumn{5}{@{}l}{\textbf{\textsc{Sentence-level NoReC (16-shot)}}} \\
\toprule
\textbf{Model} & \textbf{Vocabulary} & \textbf{Note} & \textbf{Average length} & \textbf{Time / sample} & \textbf{Slowdown} \\ \midrule
NorMistral-11B & 51\,200 & \footnotesize \textit{our new Norwegian tokenizer} & 522 tokens & 0.23 s & 1$\times$ \\

Mistral-Nemo-12B  & 131\,072 & \footnotesize \textit{original English-centric tokenizer} & 640 tokens & 0.30 s & 1.30$\times$ 
\\[1em]
NorwAI-Mistral-7B & 67\,993 & \footnotesize \textit{extends an English tokenizer} & 591 tokens & 0.18 s & 0.78$\times$ \\
NorwAI-Llama2-7B & 67\,993 & \footnotesize\textit{extends an English tokenizer} & 591 tokens & 0.15 s & 0.65$\times$\\
NorMistral-7B-warm  & 32\,768 & \footnotesize \textit{new Norwegian tokenizer} & 569 tokens & 0.17 s & 0.74$\times$ \\
NorGPT-3B & 64\,000 & \footnotesize \textit{new Norwegian tokenizer} & 552 tokens & 0.08 s & 0.35$\times$ \\
Viking-7B & 131\,072 & \footnotesize \textit{new Nordic tokenizer} & 512 tokens & 0.14 s & 0.61$\times$ \\
Viking-13B & 131\,072 & \footnotesize \textit{new Nordic tokenizer} & 512 tokens & 0.26 s & 1.13$\times$ \\
\bottomrule \\
\multicolumn{5}{@{}l}{\textbf{\textsc{NorQuAD (8-shot)}}} \\
\toprule
\textbf{Model} & \textbf{Vocabulary} & \textbf{Note} & \textbf{Average length} & \textbf{Time / sample} & \textbf{Slowdown} \\ \midrule
NorMistral-11B & 51\,200 & \footnotesize \textit{our new Norwegian tokenizer} & 4\,909 tokens & 3.10 s & 1$\times$ \\

Mistral-Nemo-12B & 131\,072 & \footnotesize \textit{original English-centric tokenizer} & 6\,171 tokens & 4.13 s & 1.33$\times$
\\[1em]
NorwAI-Mistral-7B & 67\,993 & \footnotesize\textit{extends an English tokenizer} & 5\,206 tokens & 2.28 s & 0.73$\times$ \\
NorwAI-Llama2-7B & 67\,993 & \footnotesize\textit{extends an English tokenizer} & 5\,206 tokens & 2.10 s & 0.68$\times$ \\
NorMistral-7B-warm & 32\,768 & \footnotesize \textit{new Norwegian tokenizer} & 5\,012 tokens & 2.22 s & 0.71$\times$ \\
NorGPT-3B & 64\,000 & \footnotesize \textit{new Norwegian tokenizer} & 4\,604 tokens & --- & --- \\
Viking-7B & 131\,072 & \footnotesize\textit{new Nordic tokenizer} & 4\,810 tokens & 1.93 s & 0.62$\times$ \\
Viking-13B & 131\,072 & \footnotesize\textit{new Nordic tokenizer} & 4\,810 tokens & --- & --- \\
\bottomrule
\end{tabular}%
}
\caption{\textbf{Inference speed with different tokenization strategies}\hspace{1em}We measure the average sequence length that a model needs to precess per sample, as well as the average processing time per sample. These statistics are measured on a classification task (NoReC) as well as on a generative task (NorQuAD). Some models were not able to process the dataset, either because of not supporting long-enough input sequences or because of out-of-memory errors.}
\label{tab:tokenizers-speed}
\end{table}

\section{Evaluation details}
\label{appendix:evaluation}
We provide a complete description of the evaluation design in this appendix. We provide inference details and prompts as well as full non-aggregated results here. Further information can be found at {\footnotesize\textbf{\url{https://github.com/ltgoslo/noreval}}} and {\footnotesize\textbf{\url{https://github.com/ltgoslo/norallm}}}.

\vspace{0.25em}
\subsection{Belebele}

Belebele is a reading comprehension benchmark for evaluating the natural language understanding of language models \citep{bandarkar-etal-2024-belebele}. 

\paragraph{Inference setup} The model is given a test example formatted according to a prompt template and ranks the answer candidates based on their probabilities. The most probable answer candidate is selected as the resulting answer.

\paragraph{Performance metric} There are four possible answers for each passage-question pair. We measure the performance with a simple accuracy.

\paragraph{Prompt templates} We used the following five prompt templates from \texttt{NorEval}.

{
\vspace{1em}
\noindent\footnotesize\texttt{\textbf{Prompt A:}}\vspace{-0.5em}
\begin{minted}[linenos=true, breaklines=true, baselinestretch=1.2, bgcolor=bg, breakanywhere=true, fontfamily=tt, fontsize=\footnotesize, xleftmargin=2em]{genshi}
Tekst: {$passage}
Spørsmål: {$question}
A: {$answer_1}
B: {$answer_2}
C: {$answer_3}
D: {$answer_4}
Svar: {$prediction:A/B/C/D}
\end{minted}

\noindent\footnotesize\texttt{\textbf{Prompt B:}}\vspace{-0.5em}
\begin{minted}[linenos=true, breaklines=true, baselinestretch=1.2, bgcolor=bg, breakanywhere=true, fontfamily=tt, fontsize=\footnotesize, xleftmargin=2em]{genshi}
Bakgrunn: {$passage}
Spørsmål: {$question}
Svaralternativer:
- {$answer_1}
- {$answer_2}
- {$answer_3}
- {$answer_4}
Svar: {$prediction:{$answer_1}/{$answer_2}/{$answer_3}/{$answer_4}}
\end{minted}

\noindent\footnotesize\texttt{\textbf{Prompt C:}}\vspace{-0.5em}
\begin{minted}[linenos=true, breaklines=true, baselinestretch=1.2, bgcolor=bg, breakanywhere=true, fontfamily=tt, fontsize=\footnotesize, xleftmargin=2em]{genshi}
{$question}
Hvilket av følgende mulige svar er det riktige?
A: {$answer_1}
B: {$answer_2}
C: {$answer_3}
D: {$answer_4}
Svar: {$prediction:A/B/C/D}
\end{minted}

\noindent\footnotesize\texttt{\textbf{Prompt D:}}\vspace{-0.5em}
\begin{minted}[linenos=true, breaklines=true, baselinestretch=1.2, bgcolor=bg, breakanywhere=true, fontfamily=tt, fontsize=\footnotesize, xleftmargin=2em]{genshi}
Svar på følgende spørsmål: {$question}
Svaret skal baseres på følgende tekst:
{$passage}
Velg et svar fra denne listen:
- {$answer_1}
- {$answer_2}
- {$answer_3}
- {$answer_4}
Svar: {$prediction:{$answer_1}/{$answer_2}/{$answer_3}/{$answer_4}}
\end{minted}

\noindent\footnotesize\texttt{\textbf{Prompt E:}}\vspace{-0.5em}
\begin{minted}[linenos=true, breaklines=true, baselinestretch=1.2, bgcolor=bg, breakanywhere=true, fontfamily=tt, fontsize=\footnotesize, xleftmargin=2em]{genshi}
{$passage}

{$question}

A: {$answer_1}
B: {$answer_2}
C: {$answer_3}
D: {$answer_4}

Er det riktige svaret A, B, C, eller D? {$prediction:A/B/C/D}
\end{minted}
}

\paragraph{Full results} The complete evaluation results on Belebele (Bokmål) are given in \Cref{tab:belebele-results-and-prompts}. Note that the random-guessing baseline on this task achieves accuracy of $25\%$.

\begin{table*}[h!]
\centering
\footnotesize
\begin{tabular}{@{}l@{\hspace{1em}}*{5}{c}@{}}
\toprule
& 
\multicolumn{5}{c}{\textbf{0-shot}}
\\
Prompt template & A & B & C & D & E \\
\midrule
NorMistral-11B & 45.2 & 56.7 & 32.6 & \textbf{31.1} & 22.8 \\
\\[-0.5em]
NorwAI-Mistral-7B & 29.6 & 33.4 & 27.2 & 24.8 & 22.9 \\
NorwAI-Llama2-7B & 29.6 & 38.0 & 26.4 & 25.9 & 21.2 \\
NorMistral-7B-warm & 22.9 & 37.4 & 23.2 & 27.0 & 23.0 \\
NorGPT-3B & 22.2 & 26.8 & 22.9 & 25.7 & 22.9 \\
Viking-7B & 23.8 & 27.6 & 25.4 & 26.1 & 22.8 \\
Viking-13B & 27.3 & 27.3 & 28.2 & 25.1 & 22.8 \\
Mistral-Nemo-12B & \textbf{60.6} & \cellcolor{bestres}\textbf{62.8} & \textbf{38.1} & 28.4 & \textbf{27.0} \\
\bottomrule
\end{tabular}

\caption{\textbf{Complete results on Belebele question answering (Bokmål)}\hspace{1em}We show the detailed results for each evaluated model and prompt template. The best results for each column are boldfaced, the overall best result is highlighted in blue.}
\label{tab:belebele-results-and-prompts}
\end{table*}

\vspace{0.25em}
\subsection{NorQuAD}

The second benchmark for reading comprehension, NorQuAD by \newcite{ivanova-etal-2023-norquad}, follows the scheme of extractive question-answering from SQuAD \citep{rajpurkar-etal-2016-squad}.

\paragraph{Inference setup} The model is given a test example formatted according to a prompt template and generates an answer via the greedy-search decoding strategy.

\paragraph{Performance metrics} The performance metrics are exact match (the percentage of predictions that exactly match the gold answer) and F\textsubscript{1}-score (the average N-gram overlap between the prediction and the gold answer treated as bag-of-words). 

\paragraph{Prompt templates} We used the following five prompt templates from \texttt{NorEval}.

{
\vspace{1em}
\noindent\footnotesize\texttt{\textbf{Prompt A:}}\vspace{-0.5em}
\begin{minted}[linenos=true, breaklines=true, baselinestretch=1.2, bgcolor=bg, breakanywhere=true, fontfamily=tt, fontsize=\footnotesize, xleftmargin=2em]{genshi}
Tittel: {$title}

Tekst: {$passage}

Spørsmål: {$question}

Svar: {$prediction}
\end{minted}

\newpage
\noindent\footnotesize\texttt{\textbf{Prompt B:}}\vspace{-0.5em}
\begin{minted}[linenos=true, breaklines=true, baselinestretch=1.2, bgcolor=bg, breakanywhere=true, fontfamily=tt, fontsize=\footnotesize, xleftmargin=2em]{genshi}
Tittel: {$title}

Tekst: {$passage}

Gitt teksten over, hva er svaret på følgende spørsmål? "{$question}"

Svar: {$prediction}
\end{minted}

\noindent\footnotesize\texttt{\textbf{Prompt C:}}\vspace{-0.5em}
\begin{minted}[linenos=true, breaklines=true, baselinestretch=1.2, bgcolor=bg, breakanywhere=true, fontfamily=tt, fontsize=\footnotesize, xleftmargin=2em]{genshi}
Tittel: {$title}

Tekst: {$passage}

Svar på følgende: {$question}

Svar: {$prediction}
\end{minted}

\noindent\footnotesize\texttt{\textbf{Prompt D:}}\vspace{-0.5em}
\begin{minted}[linenos=true, breaklines=true, baselinestretch=1.2, bgcolor=bg, breakanywhere=true, fontfamily=tt, fontsize=\footnotesize, xleftmargin=2em]{genshi}
Tittel: {$title}

Tekst: {$passage}

Hvordan kan man svare på spørsmålet "{$question}", gitt teksten over?

Svar: {$prediction}
\end{minted}

\noindent\footnotesize\texttt{\textbf{Prompt E:}}\vspace{-0.5em}
\begin{minted}[linenos=true, breaklines=true, baselinestretch=1.2, bgcolor=bg, breakanywhere=true, fontfamily=tt, fontsize=\footnotesize, xleftmargin=2em]{genshi}
Tittel: {$title}

Tekst: {$passage}

Gitt teksten over, besvar følgende spørsmål: "{$question}"

Svar: {$prediction}
\end{minted}
}

\paragraph{Full results} The complete evaluation results on NorQuAD (Bokmål) can be found in \autoref{tab:norquad-results-and-prompts}, both F\textsubscript{1} scores and exact-match accuracies.

\begin{table*}[!th]
\centering
\footnotesize
\begin{tabular}{@{}l@{\hspace{1em}}*{11}{c}@{}}
\textsc{\textbf{F\textsubscript{1} score}} \\
\toprule
& 
\multicolumn{5}{c}{\textbf{0-shot}}
& &
\multicolumn{5}{c}{\textbf{1-shot}}
\\
\textbf{Prompt template} & A & B & C & D & E & & A & B & C & D & E \\
\midrule
NorMistral-11B & \textbf{35.4} & 32.8 & \cellcolor{bestres}\textbf{37.9} & 16.5 & \textbf{31.8} & & \textbf{54.4} & 55.3 & \textbf{53.0} & 50.6 & 52.8 \\
\\[-0.5em]
NorwAI-Mistral-7B & 28.4 & 22.0 & 28.6 & 8.5 & 21.6 & & 41.3 & 40.7 & 41.5 & 37.7 & 42.6 \\
NorwAI-Llama2-7B & 23.1 & 18.0 & 24.2 & 7.4 & 16.7 & & 34.5 & 39.2 & 36.4 & 35.2 & 37.7 \\
NorMistral-7B-warm & 24.8 & 21.0 & 23.7 & 3.2 & 17.6 & & 37.1 & 41.9 & 40.7 & 36.0 & 41.3 \\
NorGPT-3B & 1.1 & 1.1 & 0.2 & 0.4 & 0.6 & & 0.0 & 0.0 & 0.0 & 0.0 & 0.0 \\
Viking-7B & 15.0 & 20.3 & 16.9 & 7.6 & 20.3 & & 28.8 & 29.9 & 27.3 & 26.3 & 29.7 \\
Viking-13B & 19.1 & 22.5 & 20.8 & 11.9 & 22.5 & & 35.8 & 35.8 & 35.8 & 33.1 & 35.6 \\
Mistral-Nemo-12B & 27.3 & \textbf{34.3} & 29.2 & \textbf{17.2} & 31.6 & & 49.4 & \cellcolor{bestres}\textbf{56.4} & 49.4 & \textbf{53.8} & \textbf{53.4} \\
\bottomrule \\
\textsc{\textbf{Exact match }} \\
\toprule
& 
\multicolumn{5}{c}{\textbf{0-shot}}
& &
\multicolumn{5}{c}{\textbf{1-shot}}
\\
Prompt template & A & B & C & D & E & & A & B & C & D & E \\
\midrule
NorMistral-11B & \textbf{35.4} & 32.8 & \cellcolor{bestres}\textbf{37.9} & 16.5 & \textbf{31.8} & & \textbf{54.4} & 55.3 & \textbf{53.0} & 50.6 & 52.8 \\
\\[-0.5em]
NorwAI-Mistral-7B & 28.4 & 22.0 & 28.6 & 8.5 & 21.6 & & 41.3 & 40.7 & 41.5 & 37.7 & 42.6 \\
NorwAI-Llama2-7B & 23.1 & 18.0 & 24.2 & 7.4 & 16.7 & & 34.5 & 39.2 & 36.4 & 35.2 & 37.7 \\
NorMistral-7B-warm & 24.8 & 21.0 & 23.7 & 3.2 & 17.6 & & 37.1 & 41.9 & 40.7 & 36.0 & 41.3 \\
NorGPT-3B & 1.1 & 1.1 & 0.2 & 0.4 & 0.6 & & 0.0 & 0.0 & 0.0 & 0.0 & 0.0 \\
Viking-7B & 15.0 & 20.3 & 16.9 & 7.6 & 20.3 & & 28.8 & 29.9 & 27.3 & 26.3 & 29.7 \\
Viking-13B & 19.1 & 22.5 & 20.8 & 11.9 & 22.5 & & 35.8 & 35.8 & 35.8 & 33.1 & 35.6 \\
Mistral-Nemo-12B & 27.3 & \textbf{34.3} & 29.2 & \textbf{17.2} & 31.6 & & 49.4 & \cellcolor{bestres}\textbf{56.4} & 49.4 & \textbf{53.8} & \textbf{53.4} \\
\bottomrule
\end{tabular}
\caption{\textbf{Complete results on extractive question answering with NorQuAD}\hspace{1em}We show the detailed results for each evaluated model, few-shot setting and prompt template. The best results for each column are boldfaced, the overall best result for each few-shot setting is highlighted in blue.}
\label{tab:norquad-results-and-prompts}
\end{table*}

\vspace{0.25em}
\subsection{Sentiment analysis}

Sentiment analysis can serve as a good indicator of language understanding when evaluating language models. We use NoReC as a source of manually-annotated data for sentiment analysis \citep{velldal-etal-2018-norec}. While it offers fine-grained 6-class sentiment labels, we simplify the task to binary sentiment analysis, which works more reliably for in-context learning \citep{maehlum-etal-2024-difficult}.

\paragraph{Inference setup} The model is given a test example formatted according to a prompt template and ranks the answer candidates based on their probabilities. The most probable answer candidate is selected as the resulting answer.

\paragraph{Performance metrics} The dataset is slightly unbalanced and so we use the macro-average F\textsubscript{1}-score to assess the performance.

\newpage
\subsubsection{Sentence-level NoReC}

The converted dataset with binary sentiment labels can be found at \textbf{\footnotesize\url{https://huggingface.co/datasets/ltg/norec_sentence}}.

\paragraph{Prompt templates} We used the following five prompt templates from \texttt{NorEval}.

{
\vspace{1em}
\noindent\footnotesize\texttt{\textbf{Prompt A:}}\vspace{-0.5em}
\begin{minted}[linenos=true, breaklines=true, baselinestretch=1.2, bgcolor=bg, breakanywhere=true, fontfamily=tt, fontsize=\footnotesize, xleftmargin=2em]{genshi}
Tekst: {$text}
Sentiment: {$prediction:positiv/negativ}
\end{minted}

\noindent\footnotesize\texttt{\textbf{Prompt B:}}\vspace{-0.5em}
\begin{minted}[linenos=true, breaklines=true, baselinestretch=1.2, bgcolor=bg, breakanywhere=true, fontfamily=tt, fontsize=\footnotesize, xleftmargin=2em]{genshi}
{$text}
Er denne setningen "positiv" eller "negativ"? {$prediction:positiv/negativ}
\end{minted}

\noindent\footnotesize\texttt{\textbf{Prompt C:}}\vspace{-0.5em}
\begin{minted}[linenos=true, breaklines=true, baselinestretch=1.2, bgcolor=bg, breakanywhere=true, fontfamily=tt, fontsize=\footnotesize, xleftmargin=2em]{genshi}
{$text}
Hva slags sentiment uttrykker anmelderen? {$prediction:positiv/negativ}
\end{minted}

\noindent\footnotesize\texttt{\textbf{Prompt D:}}\vspace{-0.5em}
\begin{minted}[linenos=true, breaklines=true, baselinestretch=1.2, bgcolor=bg, breakanywhere=true, fontfamily=tt, fontsize=\footnotesize, xleftmargin=2em]{genshi}
{$text}
Er anmeldelsen "positiv" eller "negativ"? {$prediction:positiv/negativ}
\end{minted}

\noindent\footnotesize\texttt{\textbf{Prompt E:}}\vspace{-0.5em}
\begin{minted}[linenos=true, breaklines=true, baselinestretch=1.2, bgcolor=bg, breakanywhere=true, fontfamily=tt, fontsize=\footnotesize, xleftmargin=2em]{genshi}
{$text}
Er denne setningen positiv eller negativ? {$prediction:positiv/negativ}
\end{minted}
}

\paragraph{Full results} The complete evaluation results on sentence-level NoReC are given in \autoref{tab:norec_sentence_prompts_and_results}. The random-guessing baseline achieves 48.5\% on this task.

\begin{table*}[!h]
\resizebox{\textwidth}{!}{%
\begin{tabular}{@{}l@{\hspace{2em}}*{17}{c}@{}}
\toprule
& 
\multicolumn{5}{c}{\textbf{0-shot}}
& &
\multicolumn{5}{c}{\textbf{1-shot}}
& &
\multicolumn{5}{c}{\textbf{16-shot}}
\\
\textbf{Prompt template} & A & B & C & D & E & & A & B & C & D & E & & A & B & C & D & E \\
\midrule
NorMistral-11B & 73.9 & 68.8 & 68.8 & 69.1 & 68.8 & & \cellcolor{bestres}\textbf{88.5\textsuperscript{\textpm1.3}} & 65.4\textsuperscript{\textpm2.0} & 79.8\textsuperscript{\textpm1.7} & 76.3\textsuperscript{\textpm1.8} & 72.4\textsuperscript{\textpm1.9} & & \textbf{90.2\textsuperscript{\textpm1.2}} & \cellcolor{bestres}\textbf{91.8\textsuperscript{\textpm1.1}} & 90.2\textsuperscript{\textpm1.2} & \textbf{91.1\textsuperscript{\textpm1.2}} & \textbf{91.6\textsuperscript{\textpm1.2}} \\
\\[-0.5em]
NorwAI-Mistral-7B & 69.8 & 55.7 & \textbf{72.7} & 53.3 & 63.0 & & 76.0\textsuperscript{\textpm1.8} & 55.1\textsuperscript{\textpm2.1} & 81.1\textsuperscript{\textpm1.6} & 77.2\textsuperscript{\textpm1.7} & 56.1\textsuperscript{\textpm2.1} & & 89.0\textsuperscript{\textpm1.3} & 86.4\textsuperscript{\textpm1.4} & \textbf{90.4\textsuperscript{\textpm1.2}} & 87.3\textsuperscript{\textpm1.4} & 87.8\textsuperscript{\textpm1.4} \\
NorwAI-Llama2-7B & 67.2 & 54.9 & 69.1 & 36.7 & 58.5 & & 72.0\textsuperscript{\textpm1.9} & 64.3\textsuperscript{\textpm2.0} & 65.9\textsuperscript{\textpm2.0} & 63.6\textsuperscript{\textpm2.0} & 65.5\textsuperscript{\textpm2.0} & & 88.2\textsuperscript{\textpm1.3} & 83.5\textsuperscript{\textpm1.5} & 88.9\textsuperscript{\textpm1.3} & 82.8\textsuperscript{\textpm1.6} & 87.5\textsuperscript{\textpm1.4} \\
NorMistral-7B-warm & \cellcolor{bestres}\textbf{75.0} & 68.4 & 61.9 & 54.7 & \textbf{69.0} & & 81.6\textsuperscript{\textpm1.6} & 69.1\textsuperscript{\textpm1.9} & 74.6\textsuperscript{\textpm1.8} & 71.5\textsuperscript{\textpm1.9} & 69.3\textsuperscript{\textpm1.9} & & 86.6\textsuperscript{\textpm1.4} & 72.4\textsuperscript{\textpm1.9} & 85.9\textsuperscript{\textpm1.4} & 77.2\textsuperscript{\textpm1.7} & 72.6\textsuperscript{\textpm1.8} \\
NorGPT-3B & 72.4 & 41.2 & 47.3 & \textbf{71.4} & 67.9 & & 66.0\textsuperscript{\textpm2.0} & 61.2\textsuperscript{\textpm2.0} & 61.9\textsuperscript{\textpm2.0} & 64.3\textsuperscript{\textpm2.0} & 59.3\textsuperscript{\textpm2.0} & & 58.0\textsuperscript{\textpm2.0} & 48.9\textsuperscript{\textpm2.1} & 65.2\textsuperscript{\textpm2.0} & 48.7\textsuperscript{\textpm2.1} & 51.1\textsuperscript{\textpm2.1} \\
Viking-7B & 70.5 & 69.0 & 70.8 & 59.0 & 67.4 & & 79.4\textsuperscript{\textpm1.7} & 70.8\textsuperscript{\textpm1.9} & 74.4\textsuperscript{\textpm1.8} & 73.6\textsuperscript{\textpm1.8} & 55.6\textsuperscript{\textpm2.1} & & 81.8\textsuperscript{\textpm1.6} & 77.4\textsuperscript{\textpm1.7} & 73.8\textsuperscript{\textpm1.8} & 76.2\textsuperscript{\textpm1.8} & 82.5\textsuperscript{\textpm1.6} \\
Viking-13B & 69.1 & \textbf{69.1} & 68.1 & 50.8 & 68.1 & & 78.9\textsuperscript{\textpm1.7} & 69.0\textsuperscript{\textpm1.9} & 79.9\textsuperscript{\textpm1.7} & 71.5\textsuperscript{\textpm1.9} & 69.0\textsuperscript{\textpm1.9} & & 84.0\textsuperscript{\textpm1.5} & 77.4\textsuperscript{\textpm1.7} & 83.0\textsuperscript{\textpm1.6} & 80.4\textsuperscript{\textpm1.6} & 79.2\textsuperscript{\textpm1.7} \\
Mistral-Nemo-12B & 71.9 & 68.4 & 68.8 & 68.4 & 69.0 & & 84.4\textsuperscript{\textpm1.5} & \textbf{77.4\textsuperscript{\textpm1.7}} & \textbf{84.6\textsuperscript{\textpm1.5}} & \textbf{82.0\textsuperscript{\textpm1.6}} & \textbf{80.3\textsuperscript{\textpm1.6}} & & 87.0\textsuperscript{\textpm1.4} & 89.0\textsuperscript{\textpm1.3} & 88.2\textsuperscript{\textpm1.3} & 87.5\textsuperscript{\textpm1.4} & 88.9\textsuperscript{\textpm1.3} \\
\bottomrule
\end{tabular}
}
\caption{\textbf{Complete results on sentence-level sentiment analysis with NoReC}\hspace{1em}We show the detailed results for each evaluated model, few-shot setting and prompt template. As the few-shot demonstrations are sampled randomly, we repeat them five times and show the mean accuracy as well as the standard deviation (rendered as superscript). The best results for each column are boldfaced, the overall best result for each few-shot setting is highlighted in blue.}
\label{tab:norec_sentence_prompts_and_results}
\end{table*}

\subsubsection{Document-level NoReC}

The converted dataset with binary sentiment labels can be found at \textbf{\footnotesize\url{https://huggingface.co/datasets/ltg/norec_document}}.

\paragraph{Prompt templates} We used the following five prompt templates from \texttt{NorEval} for testing all language models on document-level sentiment analysis:

{
\vspace{1em}
\noindent\footnotesize\texttt{\textbf{Prompt A:}}\vspace{-0.5em}
\begin{minted}[linenos=true, breaklines=true, baselinestretch=1.2, bgcolor=bg, breakanywhere=true, fontfamily=tt, fontsize=\footnotesize, xleftmargin=2em]{genshi}
Tekst: {$text}
Sentiment: {$prediction:positiv/negativ}
\end{minted}

\noindent\footnotesize\texttt{\textbf{Prompt B:}}\vspace{-0.5em}
\begin{minted}[linenos=true, breaklines=true, baselinestretch=1.2, bgcolor=bg, breakanywhere=true, fontfamily=tt, fontsize=\footnotesize, xleftmargin=2em]{genshi}
Tekst: {$text}
Er anmeldelsen "positiv" eller "negativ"? {$prediction:positiv/negativ}
\end{minted}

\noindent\footnotesize\texttt{\textbf{Prompt C:}}\vspace{-0.5em}
\begin{minted}[linenos=true, breaklines=true, baselinestretch=1.2, bgcolor=bg, breakanywhere=true, fontfamily=tt, fontsize=\footnotesize, xleftmargin=2em]{genshi}
Er polariteten til følgende anmeldelse positiv eller negativ?
Anmeldelse: {$text}
Anmeldelsen er {$prediction:positiv/negativ}
\end{minted}

\noindent\footnotesize\texttt{\textbf{Prompt D:}}\vspace{-0.5em}
\begin{minted}[linenos=true, breaklines=true, baselinestretch=1.2, bgcolor=bg, breakanywhere=true, fontfamily=tt, fontsize=\footnotesize, xleftmargin=2em]{genshi}
Anmeldelse: {$text}
Er anmelderen positiv eller negativ? {$prediction:positiv/negativ}
\end{minted}

\noindent\footnotesize\texttt{\textbf{Prompt E:}}\vspace{-0.5em}
\begin{minted}[linenos=true, breaklines=true, baselinestretch=1.2, bgcolor=bg, breakanywhere=true, fontfamily=tt, fontsize=\footnotesize, xleftmargin=2em]{genshi}
Anmeldelse: {$text}
Vil du oppsummere anmeldelsen som "bra" eller "dårlig"? {$prediction:bra/dårlig}
\end{minted}
}

\newpage
\paragraph{Full results} The complete evaluation results on document-level NoReC are provided in \autoref{tab:norec_doc_prompts_and_results}. The random-guessing baseline achieves 48.4\% on this task.

\begin{table*}[!th]
\centering
\footnotesize
\begin{tabular}{@{}l@{\hspace{2em}}*{11}{c}@{}}
\toprule
& 
\multicolumn{5}{c}{\textbf{0-shot}}
& &
\multicolumn{5}{c}{\textbf{1-shot}}
\\
\textbf{Prompt template} & A & B & C & D & E & & A & B & C & D & E \\
\midrule
NorMistral-11B & 69.4 & 67.0 & 66.6 & \textbf{68.9} & 67.0 & & 88.4\textsuperscript{\textpm0.6} & \textbf{92.5\textsuperscript{\textpm0.5}} & 86.0\textsuperscript{\textpm0.6} & \cellcolor{bestres}\textbf{92.5\textsuperscript{\textpm0.5}} & 87.7\textsuperscript{\textpm0.6} \\
\\[-0.5em]
NorwAI-Mistral-7B & 69.1 & 66.1 & 67.0 & 66.5 & 63.8 & & 82.8\textsuperscript{\textpm0.7} & 70.4\textsuperscript{\textpm0.8} & 84.6\textsuperscript{\textpm0.7} & 77.1\textsuperscript{\textpm0.8} & 84.6\textsuperscript{\textpm0.7} \\
NorwAI-Llama2-7B & 71.9 & 39.8 & 67.0 & 65.2 & 65.2 & & 76.2\textsuperscript{\textpm0.8} & 73.2\textsuperscript{\textpm0.8} & 78.7\textsuperscript{\textpm0.8} & 83.8\textsuperscript{\textpm0.7} & 80.1\textsuperscript{\textpm0.7} \\
NorMistral-7B-warm & 74.8 & 55.6 & 67.2 & 67.4 & \textbf{67.6} & & 84.3\textsuperscript{\textpm0.7} & 73.9\textsuperscript{\textpm0.8} & 84.3\textsuperscript{\textpm0.7} & 75.7\textsuperscript{\textpm0.8} & 73.4\textsuperscript{\textpm0.8} \\
NorGPT-3B & 67.7 & 52.4 & 67.0 & 67.7 & 67.0 & & 58.1\textsuperscript{\textpm0.9} & 54.0\textsuperscript{\textpm0.9} & 55.5\textsuperscript{\textpm0.9} & 54.8\textsuperscript{\textpm0.9} & 55.0\textsuperscript{\textpm0.9} \\
Viking-7B & 75.3 & 56.6 & \textbf{68.3} & 65.9 & 67.0 & & 84.5\textsuperscript{\textpm0.7} & 78.4\textsuperscript{\textpm0.8} & 73.9\textsuperscript{\textpm0.8} & 74.5\textsuperscript{\textpm0.8} & 73.4\textsuperscript{\textpm0.8} \\
Viking-13B & 69.0 & 66.8 & 67.3 & 68.3 & 65.0 & & 83.2\textsuperscript{\textpm0.7} & 72.5\textsuperscript{\textpm0.8} & 89.2\textsuperscript{\textpm0.6} & 84.5\textsuperscript{\textpm0.7} & 83.2\textsuperscript{\textpm0.7} \\
Mistral-Nemo-12B & \cellcolor{bestres}\textbf{78.5} & \textbf{68.0} & 67.0 & 67.1 & 67.0 & & \textbf{91.2\textsuperscript{\textpm0.5}} & 89.8\textsuperscript{\textpm0.6} & \textbf{90.5\textsuperscript{\textpm0.5}} & 89.8\textsuperscript{\textpm0.6} & \textbf{89.3\textsuperscript{\textpm0.6}} \\
\bottomrule
\end{tabular}
\caption{\textbf{Complete results on document-level sentiment analysis with NoReC}\hspace{1em}We show the detailed results for each evaluated model, few-shot setting and prompt template. As the few-shot demonstrations are sampled randomly, we repeat them five times and show the mean accuracy as well as the standard deviation (rendered as superscript). The best results for each column are boldfaced, the overall best result for each few-shot setting is highlighted in blue.}
\label{tab:norec_doc_prompts_and_results}
\end{table*}

\vspace{0.25em}
\subsection{NorCommonsenseQA}

Accurately predicting the correct answers on this datasets requires different types of commonsense knowledge. The creating of the Norwegian NorCommonsenseQA has been inspired by the English CommonsenseQA dataset \citep{talmor-etal-2019-commonsenseqa}. The data can be found at \textbf{\footnotesize\url{https://huggingface.co/datasets/ltg/norcommonsenseqa}}.

\paragraph{Inference setup} The model is given a test example formatted according to a prompt template and ranks the answer candidates based on their probabilities. The most probable answer candidate is selected as the resulting answer.

\paragraph{Performance metric} There are five possible answers for each question. We measure the performance with a simple accuracy.

\paragraph{Prompt templates} We used the following five prompt templates from \texttt{NorEval}. The templates are adapted to the Bokmål and Nynorsk versions of this dataset. 

{
\vspace{1em}
\noindent\footnotesize\texttt{\textbf{Prompt A (Bokmål and Nynorsk):}}\vspace{-0.5em}
\begin{minted}[linenos=true, breaklines=true, baselinestretch=1.2, bgcolor=bg, breakanywhere=true, fontfamily=tt, fontsize=\footnotesize, xleftmargin=2em]{genshi}
Spørsmål: {$question}

Svar: {$prediction:{$answer_1}/{$answer_2}/{$answer_3}/{$answer_4}/{$answer_5}}
\end{minted}

\vspace{1em}

\noindent\footnotesize\texttt{\textbf{Prompt B (Bokmål):}}\vspace{-0.5em}
\begin{minted}[linenos=true, breaklines=true, baselinestretch=1.2, bgcolor=bg, breakanywhere=true, fontfamily=tt, fontsize=\footnotesize, xleftmargin=2em]{genshi}
{$question}
Hvilket av følgende mulige svar er det riktige?
A: {$answer_1}
B: {$answer_2}
C: {$answer_3}
D: {$answer_4}
E: {$answer_5}
Svar: {$prediction:A/B/C/D/E}
\end{minted}

\newpage
\noindent\footnotesize\texttt{\textbf{Prompt B (Nynorsk):}}\vspace{-0.5em}
\begin{minted}[linenos=true, breaklines=true, baselinestretch=1.2, bgcolor=bg, breakanywhere=true, fontfamily=tt, fontsize=\footnotesize, xleftmargin=2em]{genshi}
{$question}
Kva av følgande moglege svar er det rette?
A: {$answer_1}
B: {$answer_2}
C: {$answer_3}
D: {$answer_4}
E: {$answer_5}
Svar: {$prediction:A/B/C/D/E}
\end{minted}

\vspace{1em}

\noindent\footnotesize\texttt{\textbf{Prompt C (Bokmål):}}\vspace{-0.5em}
\begin{minted}[linenos=true, breaklines=true, baselinestretch=1.2, bgcolor=bg, breakanywhere=true, fontfamily=tt, fontsize=\footnotesize, xleftmargin=2em]{genshi}
Gitt alternativene under, hva er svaret på følgende spørsmål: {$question}

Alternativer:
- {$answer_1}
- {$answer_2}
- {$answer_3}
- {$answer_4}
- {$answer_5}

Svar: {$prediction:{$answer_1}/{$answer_2}/{$answer_3}/{$answer_4}/{$answer_5}}
\end{minted}

\vspace{-0.5em}

\noindent\footnotesize\texttt{\textbf{Prompt C (Nynorsk):}}\vspace{-0.5em}
\begin{minted}[linenos=true, breaklines=true, baselinestretch=1.2, bgcolor=bg, breakanywhere=true, fontfamily=tt, fontsize=\footnotesize, xleftmargin=2em]{genshi}
Gitt alternativa under, kva er svaret på følgande spørsmål: {$question}

Alternativ:
- {$answer_1}
- {$answer_2}
- {$answer_3}
- {$answer_4}
- {$answer_5}

Svar: {$prediction:A/B/C/D/E}
\end{minted}

\vspace{1em}

\noindent\footnotesize\texttt{\textbf{Prompt D (Bokmål):}}\vspace{-0.5em}
\begin{minted}[linenos=true, breaklines=true, baselinestretch=1.2, bgcolor=bg, breakanywhere=true, fontfamily=tt, fontsize=\footnotesize, xleftmargin=2em]{genshi}
{$question}
Velg riktig svar blant disse alternativene:
- {$answer_1}
- {$answer_2}
- {$answer_3}
- {$answer_4}
- {$answer_5}

Svar: {$prediction:{$answer_1}/{$answer_2}/{$answer_3}/{$answer_4}/{$answer_5}}
\end{minted}

\vspace{-0.5em}

\noindent\footnotesize\texttt{\textbf{Prompt D (Nynorsk):}}\vspace{-0.5em}
\begin{minted}[linenos=true, breaklines=true, baselinestretch=1.2, bgcolor=bg, breakanywhere=true, fontfamily=tt, fontsize=\footnotesize, xleftmargin=2em]{genshi}
{$question}
Vel rett svar blant desse alternativa:
- {$answer_1}
- {$answer_2}
- {$answer_3}
- {$answer_4}
- {$answer_5}

Svar: {$prediction:{$answer_1}/{$answer_2}/{$answer_3}/{$answer_4}/{$answer_5}}
\end{minted}

\vspace{1em}

\noindent\footnotesize\texttt{\textbf{Prompt E (Bokmål):}}\vspace{-0.5em}
\begin{minted}[linenos=true, breaklines=true, baselinestretch=1.2, bgcolor=bg, breakanywhere=true, fontfamily=tt, fontsize=\footnotesize, xleftmargin=2em]{genshi}
{$question}
A: {$answer_1}
B: {$answer_2}
C: {$answer_3}
D: {$answer_4}
E: {$answer_5}

Er det riktige svaret A, B, C, D, eller E?

Svar: {$prediction:A/B/C/D/E}
\end{minted}

\vspace{-0.5em}

\noindent\footnotesize\texttt{\textbf{Prompt E (Nynorsk):}}\vspace{-0.5em}
\begin{minted}[linenos=true, breaklines=true, baselinestretch=1.2, bgcolor=bg, breakanywhere=true, fontfamily=tt, fontsize=\footnotesize, xleftmargin=2em]{genshi}
{$question}
A: {$answer_1}
B: {$answer_2}
C: {$answer_3}
D: {$answer_4}
E: {$answer_5}

Er det rette svaret A, B, C, D, eller E?

Svar: {$prediction:A/B/C/D/E}
\end{minted}
}

\paragraph{Full results} The complete evaluation results on NorCommonsenseQA (Bokmål and Nynorsk) are provided in \autoref{tab:csqa_prompts_and_results}. For reference, the random-guessing baseline achieves 20\% on this task (for both language variants).

\begin{table*}[!th]
\centering
\footnotesize
\begin{tabular}{@{}l@{\hspace{2em}}*{5}{c}cc*{5}{c}@{}}
\toprule
& \multicolumn{5}{c}{\textbf{Bokmål (0-shot)}} & & & \multicolumn{5}{c}{\textbf{Nynorsk (0-shot)}} \\
\cmidrule(lr){2-6} \cmidrule(lr){9-13}
Prompt template & A & B & C & D & E & & & A & B & C & D & E \\
\midrule
NorMistral-11B & \cellcolor{bestres}\textbf{61.0}& \textbf{56.9} & 23.1 & \textbf{51.8} & \textbf{45.3} & & & \textbf{44.2} & \cellcolor{bestres}\textbf{51.6} & \textbf{36.8} & \textbf{46.3} & \textbf{30.5} \\
\\[-0.5em]
NorwAI-Mistral-7B & 30.8 & 49.7 & 20.3 & 22.3 & 28.5 & & & 43.2 & 20.0 & 23.2 & 27.4 & 23.2 \\
NorwAI-Llama2-7B & 37.2 & 54.2 & 23.2 & 25.8 & 33.6 & & & 37.9 & 18.9 & 17.9 & 27.4 & 18.9 \\
NorMistral-7B-warm & 30.4 & 51.3 & 20.5 & 21.4 & 29.2 & & & 43.2 & 18.9 & 15.8 & 30.5 & 20.0 \\
NorGPT-3B & 26.4 & 34.7 & 22.1 & 20.1 & 23.5 & & & 29.5 & 20.0 & 16.8 & 25.3 & 25.3 \\
Viking-7B & 26.1 & 44.9 & 19.1 & 20.5 & 23.2 & & & 38.9 & 21.1 & 25.3 & 23.2 & 23.2 \\
Viking-13B & 24.7 & 51.1 & 18.1 & 19.1 & 24.0 & & & 40.0 & 13.7 & 24.2 & 20.0 & 16.8 \\
Mistral-Nemo-12B & 43.4 & 44.1 & \textbf{43.7} & 38.9 & 31.7 & & & 33.7 & 33.7 & 25.3 & 27.4 & 25.3 \\
\bottomrule
\end{tabular}
\caption{\textbf{Complete results on commonsense reasoning evaluated on NorCommonsenseQA (Bokmål and Nynorsk)}\hspace{1em}We show the detailed results for each evaluated model and prompt template. The best results for each column are boldfaced, the overall best result is highlighted in blue.}
\label{tab:csqa_prompts_and_results}
\end{table*}

\vspace{0.25em}
\subsection{NRK-Quiz-QA}

This question-answering dataset focuses on knowledge about Norway and its culture. The data can be found at \textbf{\footnotesize\url{https://huggingface.co/datasets/ltg/nrk_quiz_qa}}.

\paragraph{Inference setup} The model is given a test example formatted according to a \textbf{Prompt template} and ranks the answer candidates based on their probabilities. The most probable answer candidate is selected as the resulting answer.

\paragraph{Performance metric} There is a limited number of possible answers for each question. We measure the performance with a simple accuracy.

\paragraph{Prompt templates} We used the following five prompt templates from \texttt{NorEval} for testing all language models on question answering with NRK-Quiz-QA. Note that the examples in this dataset have a variable number of answer options, we show the prompt templates for four options as an example. The templates are adapted to the Bokmål and Nynorsk versions of this dataset. 

{
\vspace{1em}
\noindent\footnotesize\texttt{\textbf{Prompt A (Bokmål and Nynorsk):}}\vspace{-0.5em}
\begin{minted}[linenos=true, breaklines=true, baselinestretch=1.2, bgcolor=bg, breakanywhere=true, fontfamily=tt, fontsize=\footnotesize, xleftmargin=2em]{genshi}
Spørsmål: {$question}

Svar: {$prediction:{$answer_1}/{$answer_2}/{$answer_3}/{$answer_4}}
\end{minted}

\vspace{1em}

\noindent\footnotesize\texttt{\textbf{Prompt B (Bokmål):}}\vspace{-0.5em}
\begin{minted}[linenos=true, breaklines=true, baselinestretch=1.2, bgcolor=bg, breakanywhere=true, fontfamily=tt, fontsize=\footnotesize, xleftmargin=2em]{genshi}
{$question}

Svaralternativer:
- {$answer_1}
- {$answer_2}
- {$answer_3}
- {$answer_4}

Hva er riktig svar?

Svar: {$prediction:{$answer_1}/{$answer_2}/{$answer_3}/{$answer_4}}
\end{minted}

\vspace{-0.5em}

\noindent\footnotesize\texttt{\textbf{Prompt B (Nynorsk):}}\vspace{-0.5em}
\begin{minted}[linenos=true, breaklines=true, baselinestretch=1.2, bgcolor=bg, breakanywhere=true, fontfamily=tt, fontsize=\footnotesize, xleftmargin=2em]{genshi}
{$question}
{$question}

Svaralternativer:
- {$answer_1}
- {$answer_2}
- {$answer_3}
- {$answer_4}

Kva er rett svar?

Svar: {$prediction:{$answer_1}/{$answer_2}/{$answer_3}/{$answer_4}}
\end{minted}

\vspace{1em}

\noindent\footnotesize\texttt{\textbf{Prompt C (Bokmål):}}\vspace{-0.5em}
\begin{minted}[linenos=true, breaklines=true, baselinestretch=1.2, bgcolor=bg, breakanywhere=true, fontfamily=tt, fontsize=\footnotesize, xleftmargin=2em]{genshi}
{$question}
A: {$answer_1}
B: {$answer_2}
C: {$answer_3}
D: {$answer_4}

Er det riktige svaret A, B, C, eller D? 

Svar: {$prediction:A/B/C/D}
\end{minted}

\vspace{-0.5em}

\noindent\footnotesize\texttt{\textbf{Prompt C (Nynorsk):}}\vspace{-0.5em}
\begin{minted}[linenos=true, breaklines=true, baselinestretch=1.2, bgcolor=bg, breakanywhere=true, fontfamily=tt, fontsize=\footnotesize, xleftmargin=2em]{genshi}
{$question}
A: {$answer_1}
B: {$answer_2}
C: {$answer_3}
D: {$answer_4}

Er det rette svare A, B, C, eller D? 

Svar: {$prediction:A/B/C/D}
\end{minted}

\vspace{1em}

\noindent\footnotesize\texttt{\textbf{Prompt D (Bokmål and Nynorsk):}}\vspace{-0.5em}
\begin{minted}[linenos=true, breaklines=true, baselinestretch=1.2, bgcolor=bg, breakanywhere=true, fontfamily=tt, fontsize=\footnotesize, xleftmargin=2em]{genshi}
Spørsmål: {$question}
A: {$answer_1}
B: {$answer_2}
C: {$answer_3}
D: {$answer_4}

Svar: {$prediction:A/B/C/D}
\end{minted}

\vspace{1em}

\noindent\footnotesize\texttt{\textbf{Prompt E (Bokmål):}}\vspace{-0.5em}
\begin{minted}[linenos=true, breaklines=true, baselinestretch=1.2, bgcolor=bg, breakanywhere=true, fontfamily=tt, fontsize=\footnotesize, xleftmargin=2em]{genshi}
{$question}
Velg riktig svar blant disse alternativene:
- {$answer_1}
- {$answer_2}
- {$answer_3}
- {$answer_4}

Svar: {$prediction:{$answer_1}/{$answer_2}/{$answer_3}/{$answer_4}}
\end{minted}

\vspace{-0.5em}

\noindent\footnotesize\texttt{\textbf{Prompt E (Nynorsk):}}\vspace{-0.5em}
\begin{minted}[linenos=true, breaklines=true, baselinestretch=1.2, bgcolor=bg, breakanywhere=true, fontfamily=tt, fontsize=\footnotesize, xleftmargin=2em]{genshi}
{$question}
Vel rett svar blant desse alternativa:
- {$answer_1}
- {$answer_2}
- {$answer_3}
- {$answer_4}

Svar: {$prediction:{$answer_1}/{$answer_2}/{$answer_3}/{$answer_4}}
\end{minted}
}

\paragraph{Full results} The complete evaluation results on NRK-Quiz-QA (Bokmål and Nynorsk) are in \autoref{tab:nrk_prompts_and_results}. The random-guessing baseline achieves 28\% accuracy on the Bokmål version of this task and 27\% on the Nynorsk version.

\begin{table*}[!th]
\centering
\footnotesize
\begin{tabular}{@{}l@{\hspace{2em}}*{5}{c}cc*{5}{c}@{}}
\toprule
& \multicolumn{5}{c}{\textbf{Bokmål (0-shot)}} & & & \multicolumn{5}{c}{\textbf{Nynorsk (0-shot)}} \\
\cmidrule(lr){2-6} \cmidrule(lr){9-13}
Prompt template & A & B & C & D & E & & & A & B & C & D & E \\
\midrule
NorMistral-11B & \cellcolor{bestres}\textbf{63.7} & \textbf{50.5} & 38.6 & 41.1 & \textbf{50.6} & & & \cellcolor{bestres}\textbf{71.9} & \textbf{56.5} & \textbf{46.4}& 41.9 & \textbf{57.1} \\
\\[-0.5em]
NorwAI-Mistral-7B & 55.2 & 43.4 & 34.6 & 34.8 & 46.6 & & & 65.2 & 50.6 & 35.8 & 35.6 & 53.2 \\
NorwAI-Llama2-7B & 52.3 & 39.2 & 26.0 & 30.1 & 40.3 & & & 64.3 & 44.1 & 25.3 & 31.8 & 44.1  \\
NorMistral-7B-warm & 57.9 & 39.8 & 27.7 & 32.5 & 40.7 & & & 65.9 & 41.3 & 28.8 & 32.7 & 41.1 \\
NorGPT-3B & 33.1 & 28.2 & 26.3 & 26.1 & 27.9 & & & 37.3 & 29.6 & 25.0 & 24.7 & 30.5  \\
Viking-7B & 44.3 & 29.9 & 26.1 & 28.8 & 31.9 & & & 51.1 & 31.2 & 26.8 & 30.8 & 34.7 \\
Viking-13B & 51.0 & 31.8 & 27.8 & 30.2 & 31.6  & & & 54.8 & 34.5 & 28.0 & 30.2 & 31.9 \\
Mistral-Nemo-12B & 47.0 & 46.1 & \textbf{41.8} & \textbf{47.4} & 46.6 & & & 47.2 & 43.6 & 41.4 & \textbf{45.7} & 42.8  \\
\bottomrule
\end{tabular}
\caption{\textbf{Complete results on Norwegian-specific and world knowledge evaluated on NRK-Quiz-QA (Bokmål and Nynorsk)}\hspace{1em}We show the detailed results for each evaluated model and prompt template. The best results for each column are boldfaced, the overall best result is highlighted in blue.}
\label{tab:nrk_prompts_and_results}
\end{table*}

\vspace{0.25em}
\subsection{NorOpenBookQA}

Inspired by the English OpenBookQA \citep{mihaylov-etal-2018-suit}, this task follows the open book exams for testing human understanding of a subject. Correctly answering a question should require multi-step reasoning, common and commonsense knowledge, and rich text comprehension. The data can be found at \textbf{\footnotesize\url{https://huggingface.co/datasets/ltg/noropenbookqa}}.

\paragraph{Inference setup} The model is given a test example formatted according to a prompt template and ranks the answer candidates based on their probabilities. The most probable answer candidate is selected as the resulting answer.

\paragraph{Performance metric} There are four possible for answers for each passage-question pair. We measure the performance with a simple accuracy.

\paragraph{Prompt templates} We used the following five prompt templates from \texttt{NorEval} for testing all language models on question answering with NorOpenBookQA: The templates are adapted to the Bokmål and Nynorsk versions of this dataset.

{
\vspace{1em}
\noindent\footnotesize\texttt{\textbf{Prompt A (Bokmål and Nynorsk):}}\vspace{-0.5em}
\begin{minted}[linenos=true, breaklines=true, baselinestretch=1.2, bgcolor=bg, breakanywhere=true, fontfamily=tt, fontsize=\footnotesize, xleftmargin=2em]{genshi}
{$fact}
{$question} {$prediction:{$answer_1}/{$answer_2}/{$answer_3}/{$answer_4}}
\end{minted}

\vspace{1em}

\noindent\footnotesize\texttt{\textbf{Prompt B (Bokmål):}}\vspace{-0.5em}
\begin{minted}[linenos=true, breaklines=true, baselinestretch=1.2, bgcolor=bg, breakanywhere=true, fontfamily=tt, fontsize=\footnotesize, xleftmargin=2em]{genshi}
Faktatekst: {$fact}
Spørsmål til teksten: {$question}

Svaralternativer:
- {$answer_1}
- {$answer_2}
- {$answer_3}
- {$answer_4}

Hva er riktig svar? {$prediction:{$answer_1}/{$answer_2}/{$answer_3}/{$answer_4}}
\end{minted}

\vspace{-0.5em}

\noindent\footnotesize\texttt{\textbf{Prompt B (Nynorsk):}}\vspace{-0.5em}
\begin{minted}[linenos=true, breaklines=true, baselinestretch=1.2, bgcolor=bg, breakanywhere=true, fontfamily=tt, fontsize=\footnotesize, xleftmargin=2em]{genshi}
Faktatekst: {$fact}
Spørsmål til teksten: {$question}

Svaralternativer:
- {$answer_1}
- {$answer_2}
- {$answer_3}
- {$answer_4}

Kva er rett svar? {$prediction:{$answer_1}/{$answer_2}/{$answer_3}/{$answer_4}}
\end{minted}

\vspace{1em}

\noindent\footnotesize\texttt{\textbf{Prompt C (Bokmål):}}\vspace{-0.5em}
\begin{minted}[linenos=true, breaklines=true, baselinestretch=1.2, bgcolor=bg, breakanywhere=true, fontfamily=tt, fontsize=\footnotesize, xleftmargin=2em]{genshi}
{$fact}
{$question}
A: {$answer_1}
B: {$answer_2}
C: {$answer_3}
D: {$answer_4}

Er det riktige svaret A, B, C, eller D?

Svar: {$prediction:A/B/C/D}
\end{minted}

\vspace{-0.5em}

\noindent\footnotesize\texttt{\textbf{Prompt C (Nynorsk):}}\vspace{-0.5em}
\begin{minted}[linenos=true, breaklines=true, baselinestretch=1.2, bgcolor=bg, breakanywhere=true, fontfamily=tt, fontsize=\footnotesize, xleftmargin=2em]{genshi}
{$fact}
{$question}
A: {$answer_1}
B: {$answer_2}
C: {$answer_3}
D: {$answer_4}

Er det rette svare A, B, C, eller D? 

Svar: {$prediction:A/B/C/D}
\end{minted}

\vspace{1em}

\noindent\footnotesize\texttt{\textbf{Prompt D (Bokmål and Nynorsk):}}\vspace{-0.5em}
\begin{minted}[linenos=true, breaklines=true, baselinestretch=1.2, bgcolor=bg, breakanywhere=true, fontfamily=tt, fontsize=\footnotesize, xleftmargin=2em]{genshi}
Bakgrunn: {$fact}

Spørsmål: {$question}
A: {$answer_1}
B: {$answer_2}
C: {$answer_3}
D: {$answer_4}

Svar: {$prediction:A/B/C/D}
\end{minted}

\vspace{1em}

\noindent\footnotesize\texttt{\textbf{Prompt E (Bokmål):}}\vspace{-0.5em}
\begin{minted}[linenos=true, breaklines=true, baselinestretch=1.2, bgcolor=bg, breakanywhere=true, fontfamily=tt, fontsize=\footnotesize, xleftmargin=2em]{genshi}
Ta utgangspunkt i følgende fakta når du svarer på spørsmålet: {$fact}

{$question}
Velg riktig svar blant disse alternativene:
- {$answer_1}
- {$answer_2}
- {$answer_3}
- {$answer_4}

Svar: {$prediction:{$answer_1}/{$answer_2}/{$answer_3}/{$answer_4}}
\end{minted}

\vspace{-0.5em}

\noindent\footnotesize\texttt{\textbf{Prompt E (Nynorsk):}}\vspace{-0.5em}
\begin{minted}[linenos=true, breaklines=true, baselinestretch=1.2, bgcolor=bg, breakanywhere=true, fontfamily=tt, fontsize=\footnotesize, xleftmargin=2em]{genshi}
Ta utgangspunkt i følgande fakta når du svarar på spørsmålet: {$fact}

{$question}
Vel rett svar blant desse alternativa:
- {$answer_1}
- {$answer_2}
- {$answer_3}
- {$answer_4}

Svar: {$prediction:{$answer_1}/{$answer_2}/{$answer_3}/{$answer_4}}
\end{minted}
}

\paragraph{Full results} The complete evaluation on NorOpenBookQA (Bokmål and Nynorsk) is in \autoref{tab:noropenbookqa-results-and-prompts}. Note that randomly guessing the answers achieves 25\% on this task.


\begin{table*}[!th]
\resizebox{\textwidth}{!}{%
\begin{tabular}{@{}l@{\hspace{1em}}*{17}{c}@{}}
\textsc{\textbf{Bokmål}} \\
\toprule
& 
\multicolumn{5}{c}{\textbf{0-shot}} & &
\multicolumn{5}{c}{\textbf{1-shot}} & & 
\multicolumn{5}{c}{\textbf{16-shot}} \\
\textbf{Prompt template} & \textsc{\textbf{a}} & \textsc{\textbf{b}} & \textsc{\textbf{c}} & \textsc{\textbf{d}} & \textsc{\textbf{e}} & & \textsc{\textbf{a}} & \textsc{\textbf{b}} & \textsc{\textbf{c}} & \textsc{\textbf{d}} & \textsc{\textbf{e}} & & \textsc{\textbf{a}} & \textsc{\textbf{b}} & \textsc{\textbf{c}} & \textsc{\textbf{d}} & \textsc{\textbf{e}} \\
\midrule
NorMistral-11B & 44.6 & 55.7 & 54.0 & 67.8 & 65.8 & & 45.6\textsuperscript{\textpm2.9} & 74.5\textsuperscript{\textpm2.5} & 68.8\textsuperscript{\textpm2.7} & 76.2\textsuperscript{\textpm2.5} & 70.1\textsuperscript{\textpm2.7} & & \textbf{51.3\textsuperscript{\textpm2.9}} & 75.8\textsuperscript{\textpm2.5} & 77.9\textsuperscript{\textpm2.4} & 75.8\textsuperscript{\textpm2.5} & 76.5\textsuperscript{\textpm2.5} \\
\\[-0.5em]
NorwAI-Mistral-7B & \textbf{47.7} & 35.6 & 34.6 & 35.9 & 43.3 & & 46.3\textsuperscript{\textpm2.9} & 51.7\textsuperscript{\textpm2.9} & 34.9\textsuperscript{\textpm2.8} & 35.2\textsuperscript{\textpm2.8} & 50.0\textsuperscript{\textpm2.9} & & 50.0\textsuperscript{\textpm2.9} & 52.3\textsuperscript{\textpm2.9} & 40.9\textsuperscript{\textpm2.9} & 44.3\textsuperscript{\textpm2.9} & 51.7\textsuperscript{\textpm2.9} \\
NorwAI-Llama2-7B & 45.6 & 32.6 & 25.8 & 32.6 & 43.3 & & 45.0\textsuperscript{\textpm2.9} & 51.7\textsuperscript{\textpm2.9} & 33.2\textsuperscript{\textpm2.7} & 37.6\textsuperscript{\textpm2.8} & 46.3\textsuperscript{\textpm2.9} & & 45.0\textsuperscript{\textpm2.9} & 52.3\textsuperscript{\textpm2.9} & 32.2\textsuperscript{\textpm2.7} & 34.2\textsuperscript{\textpm2.8} & 49.7\textsuperscript{\textpm2.9} \\
NorMistral-7B-warm & 46.6 & 35.6 & 28.9 & 32.6 & 44.3 & & \textbf{46.6\textsuperscript{\textpm2.9}} & 51.7\textsuperscript{\textpm2.9} & 32.9\textsuperscript{\textpm2.7} & 34.2\textsuperscript{\textpm2.8} & 46.6\textsuperscript{\textpm2.9} & & 48.3\textsuperscript{\textpm2.9} & 49.0\textsuperscript{\textpm2.9} & 42.3\textsuperscript{\textpm2.9} & 40.3\textsuperscript{\textpm2.8} & 45.0\textsuperscript{\textpm2.9} \\
NorGPT-3B & 32.6 & 31.2 & 24.2 & 22.5 & 33.2 & & 28.5\textsuperscript{\textpm2.6} & 28.2\textsuperscript{\textpm2.6} & 23.8\textsuperscript{\textpm2.5} & 24.2\textsuperscript{\textpm2.5} & 28.5\textsuperscript{\textpm2.6} & & 29.5\textsuperscript{\textpm2.6} & 27.9\textsuperscript{\textpm2.6} & 26.5\textsuperscript{\textpm2.6} & 26.8\textsuperscript{\textpm2.6} & 28.2\textsuperscript{\textpm2.6} \\
Viking-7B & 41.9 & 26.5 & 20.8 & 28.5 & 27.9 & & 45.3\textsuperscript{\textpm2.9} & 25.8\textsuperscript{\textpm2.5} & 26.8\textsuperscript{\textpm2.6} & 24.5\textsuperscript{\textpm2.5} & 30.9\textsuperscript{\textpm2.7} & & 48.7\textsuperscript{\textpm2.9} & 26.5\textsuperscript{\textpm2.6} & 28.2\textsuperscript{\textpm2.6} & 23.8\textsuperscript{\textpm2.5} & 31.9\textsuperscript{\textpm2.7} \\
Viking-13B & 44.6 & 27.5 & 21.1 & 25.5 & 31.9 & & 45.6\textsuperscript{\textpm2.9} & 33.2\textsuperscript{\textpm2.7} & 25.2\textsuperscript{\textpm2.5} & 27.2\textsuperscript{\textpm2.6} & 38.6\textsuperscript{\textpm2.8} & & 47.0\textsuperscript{\textpm2.9} & 38.9\textsuperscript{\textpm2.8} & 29.9\textsuperscript{\textpm2.7} & 26.2\textsuperscript{\textpm2.6} & 36.9\textsuperscript{\textpm2.8} \\
Mistral-Nemo-12B & 43.6 & \textbf{60.7} & \textbf{58.1} & \cellcolor{bestres}\textbf{71.5} & \textbf{68.5} & & 44.0\textsuperscript{\textpm2.9} & \textbf{82.6\textsuperscript{\textpm2.2}} & \textbf{82.2\textsuperscript{\textpm2.2}} & \cellcolor{bestres}\textbf{82.9\textsuperscript{\textpm2.2}} & \textbf{76.2\textsuperscript{\textpm2.5}} & & 49.7\textsuperscript{\textpm2.9} & \textbf{82.9\textsuperscript{\textpm2.2}} & \textbf{82.2\textsuperscript{\textpm2.2}} & \cellcolor{bestres}\textbf{85.9\textsuperscript{\textpm2.0}} & \textbf{80.9\textsuperscript{\textpm2.3}} \\
\bottomrule \\
\textsc{\textbf{Nynorsk}} \\
\toprule
& 
\multicolumn{5}{c}{\textbf{0-shot}}
& &
\multicolumn{5}{c}{\textbf{1-shot}}
& &
\multicolumn{5}{c}{\textbf{16-shot}}
\\
Prompt template & A & B & C & D & E & & A & B & C & D & E & & A & B & C & D & E \\
\midrule
NorMistral-11B & \textbf{33.3} & \textbf{56.7} & \textbf{56.7} & 56.7 & \cellcolor{bestres}\textbf{65.6} & & 28.9\textsuperscript{\textpm4.8} & 70.0\textsuperscript{\textpm4.9} & 68.9\textsuperscript{\textpm4.9} & 71.1\textsuperscript{\textpm4.8} & 68.9\textsuperscript{\textpm4.9} & & \textbf{40.0\textsuperscript{\textpm5.2}} & 72.2\textsuperscript{\textpm4.7} & 76.7\textsuperscript{\textpm4.5} & 77.8\textsuperscript{\textpm4.4} & 77.8\textsuperscript{\textpm4.4} \\
\\[-0.5em]
NorwAI-Mistral-7B & 30.0 & 37.8 & 32.2 & 27.8 & 38.9 & & 31.1\textsuperscript{\textpm4.9} & 34.4\textsuperscript{\textpm5.0} & 28.9\textsuperscript{\textpm4.8} & 27.8\textsuperscript{\textpm4.7} & 36.7\textsuperscript{\textpm5.1} & & 37.8\textsuperscript{\textpm5.1} & 45.6\textsuperscript{\textpm5.3} & 41.1\textsuperscript{\textpm5.2} & 44.4\textsuperscript{\textpm5.3} & 42.2\textsuperscript{\textpm5.2} \\
NorwAI-Llama2-7B & 25.6 & 30.0 & 32.2 & 27.8 & 28.9 & & 28.9\textsuperscript{\textpm4.8} & 37.8\textsuperscript{\textpm5.1} & 32.2\textsuperscript{\textpm5.0} & 32.2\textsuperscript{\textpm5.0} & 36.7\textsuperscript{\textpm5.1} & & 27.8\textsuperscript{\textpm4.7} & 37.8\textsuperscript{\textpm5.1} & 32.2\textsuperscript{\textpm5.0} & 32.2\textsuperscript{\textpm5.0} & 38.9\textsuperscript{\textpm5.2} \\
NorMistral-7B-warm & 26.7 & 28.9 & 34.4 & 40.0 & 36.7 & & 27.8\textsuperscript{\textpm4.7} & 32.2\textsuperscript{\textpm5.0} & 40.0\textsuperscript{\textpm5.2} & 38.9\textsuperscript{\textpm5.2} & 43.3\textsuperscript{\textpm5.3} & & 30.0\textsuperscript{\textpm4.9} & 38.9\textsuperscript{\textpm5.2} & 41.1\textsuperscript{\textpm5.2} & 40.0\textsuperscript{\textpm5.2} & 41.1\textsuperscript{\textpm5.2} \\
NorGPT-3B & 20.0 & 27.8 & 34.4 & 30.0 & 25.6 & & 20.0\textsuperscript{\textpm4.2} & 25.6\textsuperscript{\textpm4.6} & 24.4\textsuperscript{\textpm4.6} & 23.3\textsuperscript{\textpm4.5} & 20.0\textsuperscript{\textpm4.2} & & 18.9\textsuperscript{\textpm4.1} & 23.3\textsuperscript{\textpm4.5} & 34.4\textsuperscript{\textpm5.0} & 26.7\textsuperscript{\textpm4.7} & 23.3\textsuperscript{\textpm4.5} \\
Viking-7B & 22.2 & 20.0 & 18.9 & 27.8 & 15.6 & & 30.0\textsuperscript{\textpm4.9} & 22.2\textsuperscript{\textpm4.4} & 32.2\textsuperscript{\textpm5.0} & 25.6\textsuperscript{\textpm4.6} & 27.8\textsuperscript{\textpm4.7} & & 27.8\textsuperscript{\textpm4.7} & 22.2\textsuperscript{\textpm4.4} & 27.8\textsuperscript{\textpm4.7} & 23.3\textsuperscript{\textpm4.5} & 27.8\textsuperscript{\textpm4.7} \\
Viking-13B & 30.0 & 34.4 & 17.8 & 23.3 & 31.1 & & 32.2\textsuperscript{\textpm5.0} & 27.8\textsuperscript{\textpm4.7} & 17.8\textsuperscript{\textpm4.1} & 33.3\textsuperscript{\textpm5.0} & 33.3\textsuperscript{\textpm5.0} & & 36.7\textsuperscript{\textpm5.1} & 26.7\textsuperscript{\textpm4.7} & 22.2\textsuperscript{\textpm4.4} & 18.9\textsuperscript{\textpm4.1} & 26.7\textsuperscript{\textpm4.7} \\
Mistral-Nemo-12B & 30.0 & 52.2 & 52.2 & \textbf{58.9} & 60.0 & & \textbf{35.6\textsuperscript{\textpm5.1}} & \textbf{71.1\textsuperscript{\textpm4.8}} & \textbf{77.8\textsuperscript{\textpm4.4}} & \cellcolor{bestres}\textbf{78.9\textsuperscript{\textpm4.3}} & \textbf{71.1\textsuperscript{\textpm4.8}} & & 33.3\textsuperscript{\textpm5.0} & \textbf{82.2\textsuperscript{\textpm4.1}} & \textbf{82.2\textsuperscript{\textpm4.1}} & \cellcolor{bestres}\textbf{86.7\textsuperscript{\textpm3.6}} & \textbf{81.1\textsuperscript{\textpm4.1}} \\
\bottomrule
\end{tabular}
}
\caption{\textbf{Complete results on world knowledge evaluated on NorOpenBookQA (Bokmål and Nynorsk)}\hspace{1em}We show the detailed results for each evaluated model, few-shot setting and prompt template. The best results for each column are boldfaced, the overall best result for each few-shot setting is highlighted in blue.}
\label{tab:noropenbookqa-results-and-prompts}
\end{table*}

\vspace{0.25em}
\subsection{Summarization (NorSumm)}

NorSumm by \newcite{touileb2025benchmarking} is a benchmark for abstractive summarization of Norwegian news articles. It offers another perspective on the level of Norwegian language understanding of different language models. An important feature of this dataset is that its Bokmål and Nynorsk variants are parallel.

\paragraph{Inference setup} The model is given a test example formatted according to a \textbf{Prompt template} and generates the answer via a greedy search decoding strategy.

\paragraph{Performance metric} We use Rouge-L as the standard metric for summarization \citep{lin-2004-rouge}. ROUGE-L uses longest common subsequence matching, allowing it to identify matching content even when ordered differently in generated and reference summaries.

\paragraph{Prompt templates} We used the following six prompt templates from \texttt{NorEval} for testing all language models on summarization with NorSumm. The templates are adapted to the Bokmål and Nynorsk versions of this dataset.

{
\vspace{1em}
\noindent\footnotesize\texttt{\textbf{Prompt A (Bokmål):}}\vspace{-0.5em}
\begin{minted}[linenos=true, breaklines=true, baselinestretch=1.2, bgcolor=bg, breakanywhere=true, fontfamily=tt, fontsize=\footnotesize, xleftmargin=2em]{genshi}
Skriv en oppsummering av følgende artikkel med kun noen få punkter: {$article}
Oppsummering: {$prediction}
\end{minted}

\vspace{-0.5em}

\noindent\footnotesize\texttt{\textbf{Prompt A (Nynorsk):}}\vspace{-0.5em}
\begin{minted}[linenos=true, breaklines=true, baselinestretch=1.2, bgcolor=bg, breakanywhere=true, fontfamily=tt, fontsize=\footnotesize, xleftmargin=2em]{genshi}
Skriv ei oppsummering av følgande artikkel med berre nokre få punkt: {$article}
Oppsummering: {$prediction}
\end{minted}

\vspace{1em}

\newpage
\noindent\footnotesize\texttt{\textbf{Prompt B (Bokmål):}}\vspace{-0.5em}
\begin{minted}[linenos=true, breaklines=true, baselinestretch=1.2, bgcolor=bg, breakanywhere=true, fontfamily=tt, fontsize=\footnotesize, xleftmargin=2em]{genshi}
Oppsummer følgende artikkel med noen få setninger: {$article}
Oppsummering: {$prediction}
\end{minted}

\vspace{-0.5em}

\noindent\footnotesize\texttt{\textbf{Prompt B (Nynorsk):}}\vspace{-0.5em}
\begin{minted}[linenos=true, breaklines=true, baselinestretch=1.2, bgcolor=bg, breakanywhere=true, fontfamily=tt, fontsize=\footnotesize, xleftmargin=2em]{genshi}
Oppsummer følgande artikkel med nokre få setningar: {$article}
Oppsummering: {$prediction}
\end{minted}

\vspace{1em}

\noindent\footnotesize\texttt{\textbf{Prompt C (Bokmål):}}\vspace{-0.5em}
\begin{minted}[linenos=true, breaklines=true, baselinestretch=1.2, bgcolor=bg, breakanywhere=true, fontfamily=tt, fontsize=\footnotesize, xleftmargin=2em]{genshi}
{$article}
Skriv en kort og presis oppsummering av teksten over. Språket må være klart og lett å forstå. Sørg for å ikke introdusere feil. Oppsummeringen må dekke følgende spørsmål: hvem, hva, hvor, når, og hvorfor er denne saken viktig å vite om. Oppsummeringen må være engasjerende og fremheve nøkkelinformasjon fra artikkelen. Oppsummeringen skal inneholde maksimalt 700 tegn, inkludert mellomrom. {$prediction}
\end{minted}

\vspace{-0.5em}

\noindent\footnotesize\texttt{\textbf{Prompt C (Nynorsk):}}\vspace{-0.5em}
\begin{minted}[linenos=true, breaklines=true, baselinestretch=1.2, bgcolor=bg, breakanywhere=true, fontfamily=tt, fontsize=\footnotesize, xleftmargin=2em]{genshi}
{$article}
Skriv ein kort og presis oppsummering av teksten over. Språket må vere klart og lett å forstå. Sørg for å ikkje introdusere feil. Oppsummeringa må dekkje følgande spørsmål: kven, kva, kor, når, og kvifor er denne saka viktig å vite om. Oppsummeringa må vere engasjerande og framheve nøkkelinformasjon frå artikkelen. Oppsummeringa skal innehalde maksimalt 700 tegn, inkludert mellomrom. {$prediction}
\end{minted}

\vspace{1em}

\vspace{1em}

\noindent\footnotesize\texttt{\textbf{Prompt D (Bokmål):}}\vspace{-0.5em}
\begin{minted}[linenos=true, breaklines=true, baselinestretch=1.2, bgcolor=bg, breakanywhere=true, fontfamily=tt, fontsize=\footnotesize, xleftmargin=2em]{genshi}
Gi et kortfattet sammendrag av følgende tekst: {$article} {$prediction}
\end{minted}

\vspace{-0.5em}

\noindent\footnotesize\texttt{\textbf{Prompt D (Nynorsk):}}\vspace{-0.5em}
\begin{minted}[linenos=true, breaklines=true, baselinestretch=1.2, bgcolor=bg, breakanywhere=true, fontfamily=tt, fontsize=\footnotesize, xleftmargin=2em]{genshi}
Gje eit kortfatta samandrag av følgande tekst: {$article} {$prediction}
\end{minted}

\vspace{1em}

\noindent\footnotesize\texttt{\textbf{Prompt E (Bokmål):}}\vspace{-0.5em}
\begin{minted}[linenos=true, breaklines=true, baselinestretch=1.2, bgcolor=bg, breakanywhere=true, fontfamily=tt, fontsize=\footnotesize, xleftmargin=2em]{genshi}
Lag en kort oppsummering som sammenfatter den følgende teksten i noen få punkter:
{$article}

Oppsummering: {$prediction}
\end{minted}

\vspace{-0.5em}

\noindent\footnotesize\texttt{\textbf{Prompt E (Nynorsk):}}\vspace{-0.5em}
\begin{minted}[linenos=true, breaklines=true, baselinestretch=1.2, bgcolor=bg, breakanywhere=true, fontfamily=tt, fontsize=\footnotesize, xleftmargin=2em]{genshi}
Lag ein kort oppsummering som samanfattar den følgande teksten i nokre få punkt:
{$article}

Oppsummering: {$prediction}
\end{minted}

\vspace{1em}

\noindent\footnotesize\texttt{\textbf{Prompt F (Bokmål):}}\vspace{-0.5em}
\begin{minted}[linenos=true, breaklines=true, baselinestretch=1.2, bgcolor=bg, breakanywhere=true, fontfamily=tt, fontsize=\footnotesize, xleftmargin=2em]{genshi}
Hele artikkelen:
{$article}

Hovedpunkter: {$prediction}
\end{minted}

\vspace{-0.5em}

\noindent\footnotesize\texttt{\textbf{Prompt F (Nynorsk):}}\vspace{-0.5em}
\begin{minted}[linenos=true, breaklines=true, baselinestretch=1.2, bgcolor=bg, breakanywhere=true, fontfamily=tt, fontsize=\footnotesize, xleftmargin=2em]{genshi}
Heile artikkelen:
{$article}

Hovudpunkt: {$prediction}
\end{minted}
}

\paragraph{Full results} The complete evaluation on NorSumm (both Bokmål and Nynorsk variants) is provided in \Cref{tab:norsumm_prompts_and_results}.

\begin{table}[h!]
\centering
\resizebox{0.8\textwidth}{!}{%
\begin{tabular}{@{}l@{\hspace{1em}}*{6}{c}c@{\hspace{2em}}c*{6}{c}@{}}
\toprule
& \multicolumn{6}{c}{\textbf{Bokmål (0-shot)}} & & & \multicolumn{6}{c}{\textbf{Nynorsk (0-shot)}} \\
\cmidrule(lr){2-7} \cmidrule(lr){10-15}
\textbf{Prompt template} & A & B & C & D & E & F & & & A & B & C & D & E & F \\
\midrule
NorMistral-11B & \textbf{17.6} & 20.5 & \textbf{34.9} & \cellcolor{bestres}\textbf{45.0} & 42.5 & \textbf{5.7} &&& \textbf{15.4} & 17.9 & 25.4 & \textbf{32.4} & \cellcolor{bestres}\textbf{32.6} & \textbf{6.8}   \\
\\[-0.5em]
NorwAI-Mistral-7B & 12.2 & 0.0 & 0.0 & 2.6 & 0.0 & 0.0 && & 10.3 & 0.0 & 0.0 & 3.5 & 0.0 & 0.0 \\
NorwAI-Llama2-7B & 10.7 & 0.0 & 0.0 & 7.8 & 3.5 & 0.0 && & 10.4 & 0.0 & 0.0 & 4.4 & 5.9 & 0.0  \\
NorMistral-7B-warm & 7.8 & 0.0 & 0.0 & 16.5 & 9.0 & 0.0 && & 8.6 & 0.0 & 0.0 & 6.9 & 4.1 & 0.0 \\
NorGPT-3B & 8.8 & 8.5 & 31.6 & 33.8 & 25.2 & 2.8 && & 7.4 & 10.9 & 21.6 & 24.3 & 20.0 & 4.0 \\
Viking-7B & 11.2 & 5.5 & 16.5 & 29.8 & 31.9 & 0.0 && & 10.5 & 3.0 & 16.4 & 25.7 & 24.2 & 0.0  \\
Viking-13B & 11.1 & 1.7 & 6.0 & 23.7 & 36.3 & 0.0 && & 9.8 & 0.0 & 4.3 & 19.6 & 28.8 & 0.4  \\
Mistral-Nemo-12B & 13.4 & \textbf{26.4} & 41.5 & 35.6 & \textbf{44.9} & 2.9  && & 12.4 & \textbf{18.2} & \textbf{30.0} & 30.3 & 30.9 & 3.6  \\
\bottomrule
\end{tabular}
}
\caption{\textbf{Complete results on NorSumm summarization (Bokmål and Nynorsk versions)}\hspace{1em}We show the detailed results for each evaluated model and prompt template. The best results for each column are boldfaced, the overall best result is highlighted in blue.}
\label{tab:norsumm_prompts_and_results}
\end{table}

\vspace{0.25em}
\subsection{Grammatical error correction (ASK-GEC)}

This tasks tests how do language models understand more low-level features of the Norwegian language. We use the ASK-GEC dataset from \newcite{jentoft2023grammatical} that is based on corrected essays of Norwegian language learners.

\paragraph{Inference setup} The model is given a test example formatted according to a prompt template; given this input, it then generates the answer via a greedy-search decoding strategy.

\paragraph{Performance metric} We use the F\textsubscript{0.5}-score to measure the amount of successfully fixed correction-spans. These spans are heuristically identified by the ERRANT system \citep{bryant-etal-2017-automatic}. More details about using this metric for Norwegian grammatical error corrections can be found in \newcite{jentoft2023grammatical}.

\paragraph{Prompt templates} We used the following five prompt templates for grammatical error correction:

{
\vspace{1em}
\noindent\footnotesize\texttt{\textbf{Prompt A:}}\vspace{-0.5em}
\begin{minted}[linenos=true, breaklines=true, baselinestretch=1.2, bgcolor=bg, breakanywhere=true, fontfamily=tt, fontsize=\footnotesize, xleftmargin=2em]{genshi}
Tekst: {$text}
Korreksjon: {$prediction}
\end{minted}

\noindent\footnotesize\texttt{\textbf{Prompt B:}}\vspace{-0.5em}
\begin{minted}[linenos=true, breaklines=true, baselinestretch=1.2, bgcolor=bg, breakanywhere=true, fontfamily=tt, fontsize=\footnotesize, xleftmargin=2em]{genshi}
Tekst: {$text}
Rettet versjon: {$prediction}
\end{minted}

\noindent\footnotesize\texttt{\textbf{Prompt C:}}\vspace{-0.5em}
\begin{minted}[linenos=true, breaklines=true, baselinestretch=1.2, bgcolor=bg, breakanywhere=true, fontfamily=tt, fontsize=\footnotesize, xleftmargin=2em]{genshi}
Skriv om følgende tekst slik at den blir grammatisk korrekt: {$text}
Korreksjon: {$prediction}
\end{minted}

\noindent\footnotesize\texttt{\textbf{Prompt D:}}\vspace{-0.5em}
\begin{minted}[linenos=true, breaklines=true, baselinestretch=1.2, bgcolor=bg, breakanywhere=true, fontfamily=tt, fontsize=\footnotesize, xleftmargin=2em]{genshi}
Original versjon: {$text}
Korrekturlest og rettet versjon: {$prediction}
\end{minted}

\noindent\footnotesize\texttt{\textbf{Prompt E:}}\vspace{-0.5em}
\begin{minted}[linenos=true, breaklines=true, baselinestretch=1.2, bgcolor=bg, breakanywhere=true, fontfamily=tt, fontsize=\footnotesize, xleftmargin=2em]{genshi}
Rett opp grammatiske feil i denne teksten: {$text}
Korreksjon: {$prediction}
\end{minted}
}

\paragraph{Full results} The complete evaluation on ASK-GEC is provided in \Cref{tab:gec-results-by-shots-and-prompts}.

\begin{table*}[!h]
\resizebox{\textwidth}{!}{%
\begin{tabular}{@{}l@{\hspace{1em}}*{17}{c}@{}}
\toprule
& 
\multicolumn{5}{c}{\textbf{0-shot}}
& &
\multicolumn{5}{c}{\textbf{1-shot}}
& &
\multicolumn{5}{c}{\textbf{16-shot}}
\\
Prompt template & A & B & C & D & E & & A & B & C & D & E & & A & B & C & D & E \\
\midrule
NorMistral-11B & 2.8 & 16.6 & \cellcolor{bestres}\textbf{39.9} & 26.8 & \textbf{38.8} & & 38.6 & 41.8 & 43.3 & \textbf{45.2} & 44.7 & & \textbf{52.6} & 52.3 & 50.4 & 51.5 & 51.4 \\
\\[-0.5em]
NorwAI-Mistral-7B & 0.0 & 0.0 & 22.2 & 0.0 & 0.0 & & 37.2 & 39.1 & 45.3 & 42.8 & \textbf{46.1} & & 51.9 & \textbf{52.4} & \textbf{52.5} & \cellcolor{bestres}\textbf{53.2} & \textbf{52.7} \\
NorwAI-Llama2-7B & 0.0 & 0.2 & 13.3 & 0.0 & 27.1 & & 38.5 & 40.5 & \cellcolor{bestres}\textbf{46.2} & 44.2 & 45.0 & & 51.1 & 51.3 & 51.2 & 51.4 & 51.1 \\
NorMistral-7B-warm & 2.1 & 0.0 & 11.1 & \textbf{33.6} & 18.7 & & 34.4 & 38.0 & 42.8 & 41.2 & 41.1 & & 48.0 & 48.2 & 48.7 & 48.2 & 48.5 \\
NorGPT-3B & 0.2 & 0.0 & 0.2 & 1.4 & 0.3 & & 0.4 & 0.4 & 0.4 & 0.4 & 0.4 & & 0.9 & 1.3 & 1.8 & 1.2 & 1.5 \\
Viking-7B & 2.8 & 1.2 & 23.1 & 0.0 & 11.4 & & 29.9 & 37.0 & 40.6 & 39.9 & 38.9 & & 50.7 & 51.0 & 50.4 & 51.2 & 50.1 \\
Viking-13B & 3.1 & 0.0 & 37.8 & 25.1 & 34.8 & & \textbf{42.6} & \textbf{43.5} & 45.7 & 44.8 & 46.0 & & 52.4 & 52.0 & 51.9 & 52.4 & 51.8 \\
Mistral-Nemo-12B & \textbf{14.7} & \textbf{18.6} & 36.5 & 16.9 & 12.3 & & 38.8 & 36.8 & 37.5 & 38.6 & 39.6 & & 43.9 & 43.7 & 42.7 & 43.7 & 43.1 \\
\bottomrule
\end{tabular}
}
\caption{\textbf{Complete results on grammatical error correction}\hspace{1em}We show the detailed results for each evaluated model, few-shot setting and prompt template. The best results for each column are boldfaced, the overall best result for each few-shot setting is highlighted in blue.}
\end{table*}

\vspace{0.25em}
\subsection{Language identification (SLIDE)}

We use the Scandinavian language identification and evaluation (SLIDE) from \textbf{\footnotesize\url{https://github.com/ltgoslo/slide}}. This dataset consists of sentences manually annotated with the language they are written in: Norwegian Bokmål, Nynorsk, Danish or Swedish (we filtered out the examples that are not written in any Scandinavian language). The sentences can be annotated with multiple language labels if applicable.

\paragraph{Inference setup} This task is solved as classification -- the label with the highest probability given the prompt (estimated by the evaluated language model) is chosen as the predicted label. The few-shot demonstrations are randomly sampled from the SLIDE validation set.

\paragraph{Performance metric} We test whether a language model is able to correctly predict one of the (potentially multiple) languages a sentence is written in. We thus adopt the \textit{loose accuracy} metric from SLIDE, where a single-label prediction is considered to be correct if is in the set of gold language labels. 

\paragraph{Prompt templates} We used the following five prompt templates for testing all language models on grammatical error correction. The few-shot demonstrations are separated by double newlines \texttt{\textbackslash n\textbackslash n}.

{
\vspace{1em}
\noindent\footnotesize\texttt{\textbf{Prompt A:}}\vspace{-0.5em}
\begin{minted}[linenos=true, breaklines=true, baselinestretch=1.2, bgcolor=bg, breakanywhere=true, fontfamily=tt, fontsize=\footnotesize, xleftmargin=2em]{genshi}
Tekst: {$text}
Korreksjon: {$prediction}
\end{minted}

\noindent\footnotesize\texttt{\textbf{Prompt B:}}\vspace{-0.5em}
\begin{minted}[linenos=true, breaklines=true, baselinestretch=1.2, bgcolor=bg, breakanywhere=true, fontfamily=tt, fontsize=\footnotesize, xleftmargin=2em]{genshi}
Tekst: {$text}
Rettet versjon: {$prediction}
\end{minted}

\noindent\footnotesize\texttt{\textbf{Prompt C:}}\vspace{-0.5em}
\begin{minted}[linenos=true, breaklines=true, baselinestretch=1.2, bgcolor=bg, breakanywhere=true, fontfamily=tt, fontsize=\footnotesize, xleftmargin=2em]{genshi}
Skriv om følgende tekst slik at den blir grammatisk korrekt: {$text}
Korreksjon: {$prediction}
\end{minted}

\noindent\footnotesize\texttt{\textbf{Prompt D:}}\vspace{-0.5em}
\begin{minted}[linenos=true, breaklines=true, baselinestretch=1.2, bgcolor=bg, breakanywhere=true, fontfamily=tt, fontsize=\footnotesize, xleftmargin=2em]{genshi}
Original versjon: {$text}
Korrekturlest og rettet versjon: {$prediction}
\end{minted}

\newpage
\noindent\footnotesize\texttt{\textbf{Prompt E:}}\vspace{-0.5em}
\begin{minted}[linenos=true, breaklines=true, baselinestretch=1.2, bgcolor=bg, breakanywhere=true, fontfamily=tt, fontsize=\footnotesize, xleftmargin=2em]{genshi}
Rett opp grammatiske feil i denne teksten: {$text}
Korreksjon: {$prediction}
\end{minted}
}

\paragraph{Full results} The complete evaluation on language identification is given in \Cref{tab:gec-results-by-shots-and-prompts}. Note that the majority baseline on this task is $40.3\%$ loose accuracy and the random baseline is $28.2\%$. 

\begin{table*}[!th]
\resizebox{\textwidth}{!}{%
\begin{tabular}{@{}l@{\hspace{1em}}*{17}{c}@{}}
\toprule
& 
\multicolumn{5}{c}{\textbf{0-shot}} & &
\multicolumn{5}{c}{\textbf{1-shot}} & & 
\multicolumn{5}{c}{\textbf{16-shot}} \\
\textbf{Prompt template} & \textsc{\textbf{a}} & \textsc{\textbf{b}} & \textsc{\textbf{c}} & \textsc{\textbf{d}} & \textsc{\textbf{e}} & & \textsc{\textbf{a}} & \textsc{\textbf{b}} & \textsc{\textbf{c}} & \textsc{\textbf{d}} & \textsc{\textbf{e}} & & \textsc{\textbf{a}} & \textsc{\textbf{b}} & \textsc{\textbf{c}} & \textsc{\textbf{d}} & \textsc{\textbf{e}} \\
\midrule
NorMistral-11B & 74.0 & 41.5 & \textbf{55.7} & 62.6 & 53.1 & & 78.8 & \textbf{60.8\textsuperscript{\textpm0.7}} & \textbf{80.6\textsuperscript{\textpm0.2}} & \textbf{80.6\textsuperscript{\textpm0.4}} & 58.7 & & 97.9 & \textbf{95.6\textsuperscript{\textpm0.2}} & 94.6 & \textbf{97.0\textsuperscript{\textpm0.3}} & \cellcolor{bestres}\textbf{98.2\textsuperscript{\textpm0.1}} \\
\\[-0.5em]
NorwAI-Mistral-7B & 65.8 & 54.0 & 38.8 & \textbf{69.6} & \textbf{66.9} & & 74.8 & 39.2 & 47.9 & 76.9 & 64.5 & & 95.1 & 69.9 & 90.4 & 92.0 & 95.7 \\
NorwAI-Llama2-7B & 69.8 & 37.4 & 42.9 & 49.4 & 59.0 & & 65.4 & 33.0 & 40.8 & 54.5 & 43.1 & & 93.5 & 74.1 & 67.0 & 77.4 & 87.5 \\
NorMistral-7B-warm & \cellcolor{bestres}\textbf{87.5} & 47.7 & 42.2 & 65.4 & 61.6 & & \cellcolor{bestres}\textbf{85.7\textsuperscript{\textpm0.4}} & 40.3 & 63.5 & 72.2 & \textbf{73.2\textsuperscript{\textpm0.4}} & & \textbf{98.1\textsuperscript{\textpm0.1}} & 92.2 & \textbf{96.2\textsuperscript{\textpm0.3}} & 92.1 & 97.3 \\
NorGPT-3B & 36.6 & 24.0 & 49.9 & 43.9 & 42.6 & & 37.7 & 35.6 & 32.4 & 32.4 & 32.4 & & 39.0 & 27.9 & 40.0 & 40.3 & 40.2 \\
Viking-7B & 74.4 & 42.7 & 41.0 & 34.7 & 32.8 & & 46.2 & 37.1 & 35.2 & 39.5 & 36.0 & & 77.2 & 47.4 & 44.3 & 58.5 & 52.6 \\
Viking-13B & 71.5 & \textbf{59.5} & 41.0 & 41.9 & 32.4 & & 55.1 & 36.4 & 37.8 & 43.8 & 34.1 & & 84.4 & 62.0 & 56.8 & 79.1 & 65.3 \\
Mistral-Nemo-12B & 68.3 & 41.7 & 50.3 & 48.5 & 40.7 & & 63.8 & 56.0 & 74.3 & 58.6 & 45.8 & & 85.9 & 84.6 & 86.6 & 87.3 & 86.1 \\
\bottomrule
\end{tabular}
}
\caption{\textbf{Complete results on Scandinavian language identification}\hspace{1em}We show the detailed results for each evaluated model, few-shot setting and prompt template. As the few-shot demonstrations are sampled randomly, we repeat them five times and show the mean accuracy as well as the standard deviation (rendered as superscript). The best results for each column are boldfaced, the overall best result for each few-shot setting is highlighted in blue.}
\label{tab:gec-results-by-shots-and-prompts}
\end{table*}

\vspace{0.25em}
\subsection{Translation}

\paragraph{Inference setup} This task is solved as generation via prefix prompting -- the model is given a prompt without the \texttt{\small\textbf{\$prediction}} suffix and then it autoregressively generates a prediction until outputting a newline. We use simple greedy search to generate the output.

\paragraph{Performance metric} We measure the translation quality with SacreBLEU scores \citep{post-2018-call} with signature \texttt{BLEU+case.mixed+numrefs.1+smooth.exp+tok.intl+version.1.2.20} as the main metric. We also provide chrF++ scores as an additional metric \citep{popovic-2017-chrf}.

\subsubsection{English to Bokmål translation}

\paragraph{Prompt templates} We used the following four prompt templates for testing all language models on translation to Northern Sámi.

{
\vspace{1em}
\noindent\footnotesize\texttt{\textbf{Prompt A:}}\vspace{-0.5em}
\begin{minted}[linenos=true, breaklines=true, baselinestretch=1.2, bgcolor=bg, breakanywhere=true, fontfamily=tt, fontsize=\footnotesize, xleftmargin=2em]{genshi}
Engelsk: {$text}
Bokmål: {$prediction}
\end{minted}

\noindent\footnotesize\texttt{\textbf{Prompt B:}}\vspace{-0.5em}
\begin{minted}[linenos=true, breaklines=true, baselinestretch=1.2, bgcolor=bg, breakanywhere=true, fontfamily=tt, fontsize=\footnotesize, xleftmargin=2em]{genshi}
Oversett følgende setning til Bokmål: {$text}
Bokmål: {$prediction}
\end{minted}

\noindent\footnotesize\texttt{\textbf{Prompt C:}}\vspace{-0.5em}
\begin{minted}[linenos=true, breaklines=true, baselinestretch=1.2, bgcolor=bg, breakanywhere=true, fontfamily=tt, fontsize=\footnotesize, xleftmargin=2em]{genshi}
Gi en oversettelse til Bokmål for denne setningen: {$text}
Bokmål: {$prediction}
\end{minted}

\noindent\footnotesize\texttt{\textbf{Prompt D:}}\vspace{-0.5em}
\begin{minted}[linenos=true, breaklines=true, baselinestretch=1.2, bgcolor=bg, breakanywhere=true, fontfamily=tt, fontsize=\footnotesize, xleftmargin=2em]{genshi}
Hva blir "{$text}" på Bokmål?
Bokmål: {$prediction}
\end{minted}
}

\paragraph{Full results} The complete evaluation on translation to Bokmål is given in \Cref{tab:en-se-results-by-shots-and-prompts} (BLEU scores in the top sub-table and chrF++ scores below).

\begin{table*}[!th]
\resizebox{\textwidth}{!}{%
\begin{tabular}{@{}l@{\hspace{1em}}*{14}{c}@{}}
\textsc{\textbf{BLEU scores}} \\
\toprule
& 
\multicolumn{4}{c}{\textbf{0-shot}}
& &
\multicolumn{4}{c}{\textbf{1-shot}}
& &
\multicolumn{4}{c}{\textbf{16-shot}}
\\
\textbf{Prompt template} & A & B & C & D & & A & B & C & D & & A & B & C & D \\
\midrule
NorMistral-11B & 54.9 & 51.6 & 43.1 & 44.5 & & 58.2\textsuperscript{\textpm0.6} & \textbf{58.5\textsuperscript{\textpm0.6}} & 58.2\textsuperscript{\textpm0.6} & \textbf{58.1\textsuperscript{\textpm0.6}} & & 58.6\textsuperscript{\textpm0.6} & 58.8\textsuperscript{\textpm0.6} & 58.3\textsuperscript{\textpm0.6} & 58.5\textsuperscript{\textpm0.6} \\
\\[-0.5em]
NorwAI-Mistral-7B & \textbf{56.1} & 31.0 & \textbf{52.8} & 50.9 & & 58.2\textsuperscript{\textpm0.6} & 58.1\textsuperscript{\textpm0.6} & 58.1\textsuperscript{\textpm0.6} & 57.7\textsuperscript{\textpm0.7} & & 58.4\textsuperscript{\textpm0.6} & 58.7\textsuperscript{\textpm0.6} & 58.5\textsuperscript{\textpm0.6} & 58.2\textsuperscript{\textpm0.6} \\
NorwAI-Llama2-7B & 35.5 & 23.9 & 23.6 & 44.8 & & 55.8\textsuperscript{\textpm0.7} & 57.0\textsuperscript{\textpm0.6} & 57.3\textsuperscript{\textpm0.6} & 56.1\textsuperscript{\textpm0.7} & & 57.8\textsuperscript{\textpm0.6} & 57.9\textsuperscript{\textpm0.6} & 57.8\textsuperscript{\textpm0.6} & 57.5\textsuperscript{\textpm0.6} \\
NorMistral-7B-warm & 54.7 & 53.1 & 0.0 & \textbf{51.6} & & 55.9\textsuperscript{\textpm0.7} & 56.5\textsuperscript{\textpm0.6} & 56.6\textsuperscript{\textpm0.6} & 56.2\textsuperscript{\textpm0.7} & & 57.0\textsuperscript{\textpm0.7} & 57.0\textsuperscript{\textpm0.6} & 57.2\textsuperscript{\textpm0.7} & 56.4\textsuperscript{\textpm0.6} \\
NorGPT-3B & 0.1 & 0.1 & 0.1 & 0.2 & & 0.0\textsuperscript{\textpm0.0} & 0.2\textsuperscript{\textpm0.0} & 0.1\textsuperscript{\textpm0.0} & 0.4\textsuperscript{\textpm0.0} & & 0.2\textsuperscript{\textpm0.0} & 1.8\textsuperscript{\textpm0.1} & 0.7\textsuperscript{\textpm0.0} & 0.9\textsuperscript{\textpm0.0} \\
Viking-7B & 53.4 & 35.0 & 42.1 & 1.4 & & 54.2\textsuperscript{\textpm0.6} & 58.3\textsuperscript{\textpm0.7} & 57.2\textsuperscript{\textpm0.7} & 56.7\textsuperscript{\textpm0.7} & & 58.7\textsuperscript{\textpm0.7} & 59.6\textsuperscript{\textpm0.6} & 59.7\textsuperscript{\textpm0.6} & 59.4\textsuperscript{\textpm0.7} \\
Viking-13B & 24.2 & \cellcolor{bestres}\textbf{58.2} & 14.9 & 1.9 & & \textbf{58.6\textsuperscript{\textpm0.7}} & 58.3\textsuperscript{\textpm0.7} & \cellcolor{bestres}\textbf{58.7\textsuperscript{\textpm0.8}} & 57.5\textsuperscript{\textpm0.7} & & \textbf{59.5\textsuperscript{\textpm0.6}} & \cellcolor{bestres}\textbf{60.0\textsuperscript{\textpm0.6}} & \textbf{59.9\textsuperscript{\textpm0.6}} & \textbf{59.9\textsuperscript{\textpm0.7}} \\
Mistral-Nemo-12B & 44.3 & 46.1 & 44.3 & 45.5 & & 48.5\textsuperscript{\textpm0.6} & 48.9\textsuperscript{\textpm0.6} & 48.9\textsuperscript{\textpm0.6} & 48.9\textsuperscript{\textpm0.6} & & 49.3\textsuperscript{\textpm0.6} & 49.5\textsuperscript{\textpm0.6} & 49.2\textsuperscript{\textpm0.6} & 49.5\textsuperscript{\textpm0.6} \\
\bottomrule \\
\textsc{\textbf{chrF++ scores}} \\
\toprule
& 
\multicolumn{4}{c}{\textbf{0-shot}}
& &
\multicolumn{4}{c}{\textbf{1-shot}}
& &
\multicolumn{4}{c}{\textbf{16-shot}}
\\
\textbf{Prompt template} & A & B & C & D & & A & B & C & D & & A & B & C & D \\
\midrule
NorMistral-11B & \textbf{71.7} & 71.4 & 67.2 & 69.3 & & \textbf{73.6\textsuperscript{\textpm0.4}} & 73.8\textsuperscript{\textpm0.4} & \textbf{73.8\textsuperscript{\textpm0.4}} & \textbf{73.4\textsuperscript{\textpm0.4}} & & 73.8\textsuperscript{\textpm0.4} & 74.0\textsuperscript{\textpm0.4} & 73.8\textsuperscript{\textpm0.4} & 73.7\textsuperscript{\textpm0.4} \\
\\[-0.5em]
NorwAI-Mistral-7B & 71.1 & 55.5 & \textbf{68.8} & \textbf{70.8} & & 73.0\textsuperscript{\textpm0.4} & 73.2\textsuperscript{\textpm0.4} & 73.2\textsuperscript{\textpm0.4} & 72.6\textsuperscript{\textpm0.5} & & 73.4\textsuperscript{\textpm0.4} & 73.7\textsuperscript{\textpm0.4} & 73.6\textsuperscript{\textpm0.4} & 73.4\textsuperscript{\textpm0.4} \\
NorwAI-Llama2-7B & 59.3 & 40.8 & 39.9 & 64.1 & & 71.3\textsuperscript{\textpm0.4} & 72.1\textsuperscript{\textpm0.4} & 72.4\textsuperscript{\textpm0.4} & 71.2\textsuperscript{\textpm0.5} & & 72.7\textsuperscript{\textpm0.4} & 72.9\textsuperscript{\textpm0.4} & 72.9\textsuperscript{\textpm0.4} & 72.6\textsuperscript{\textpm0.4} \\
NorMistral-7B-warm & 69.6 & 67.5 & 9.6 & 66.3 & & 70.9\textsuperscript{\textpm0.4} & 71.6\textsuperscript{\textpm0.4} & 71.6\textsuperscript{\textpm0.4} & 71.3\textsuperscript{\textpm0.4} & & 72.0\textsuperscript{\textpm0.4} & 72.3\textsuperscript{\textpm0.4} & 72.2\textsuperscript{\textpm0.4} & 72.0\textsuperscript{\textpm0.4} \\
NorGPT-3B & 9.8 & 8.1 & 9.7 & 9.5 & & 4.4\textsuperscript{\textpm0.1} & 4.8\textsuperscript{\textpm0.1} & 4.4\textsuperscript{\textpm0.1} & 6.4\textsuperscript{\textpm0.1} & & 4.4\textsuperscript{\textpm0.1} & 12.3\textsuperscript{\textpm0.3} & 7.9\textsuperscript{\textpm0.1} & 8.9\textsuperscript{\textpm0.1} \\
Viking-7B & 70.8 & 51.8 & 63.6 & 12.5 & & 70.6\textsuperscript{\textpm0.4} & 72.9\textsuperscript{\textpm0.7} & 72.3\textsuperscript{\textpm0.5} & 71.9\textsuperscript{\textpm0.5} & & 73.3\textsuperscript{\textpm0.5} & 74.3\textsuperscript{\textpm0.4} & 74.4\textsuperscript{\textpm0.4} & 74.1\textsuperscript{\textpm0.5} \\
Viking-13B & 59.4 & \cellcolor{bestres}\textbf{72.8} & 46.8 & 15.5 & & 73.3\textsuperscript{\textpm0.5} & \cellcolor{bestres}\textbf{74.0\textsuperscript{\textpm0.4}} & 73.6\textsuperscript{\textpm0.5} & 72.6\textsuperscript{\textpm0.5} & & \textbf{74.2\textsuperscript{\textpm0.4}} & \cellcolor{bestres}\textbf{74.6\textsuperscript{\textpm0.4}} & \textbf{74.5\textsuperscript{\textpm0.4}} & \textbf{74.5\textsuperscript{\textpm0.4}} \\
Mistral-Nemo-12B & 61.4 & 64.2 & 61.7 & 63.4 & & 65.8\textsuperscript{\textpm0.5} & 66.5\textsuperscript{\textpm0.4} & 66.4\textsuperscript{\textpm0.4} & 66.6\textsuperscript{\textpm0.4} & & 66.8\textsuperscript{\textpm0.4} & 67.0\textsuperscript{\textpm0.4} & 66.9\textsuperscript{\textpm0.4} & 67.0\textsuperscript{\textpm0.4} \\
\bottomrule
\end{tabular}
}
\caption{\textbf{Complete results on translation from English to Norwegian Bokmål}\hspace{1em}We show the detailed results for each evaluated model, few-shot setting and prompt template. As the few-shot demonstrations are sampled randomly, we repeat them five times and show the mean accuracy as well as the standard deviation (rendered as superscript). The best results for each column are boldfaced, the overall best result for each few-shot setting is highlighted in blue.}
\end{table*}

\subsubsection{English to Nynorsk translation}

\paragraph{Prompt templates} We used the following five prompt templates for testing all language models on translation to Nynorsk.

{
\vspace{1em}
\noindent\footnotesize\texttt{\textbf{Prompt A:}}\vspace{-0.5em}
\begin{minted}[linenos=true, breaklines=true, baselinestretch=1.2, bgcolor=bg, breakanywhere=true, fontfamily=tt, fontsize=\footnotesize, xleftmargin=2em]{genshi}
Engelsk: {$text}
Nynorsk: {$prediction}
\end{minted}

\noindent\footnotesize\texttt{\textbf{Prompt B:}}\vspace{-0.5em}
\begin{minted}[linenos=true, breaklines=true, baselinestretch=1.2, bgcolor=bg, breakanywhere=true, fontfamily=tt, fontsize=\footnotesize, xleftmargin=2em]{genshi}
Omsett følgande setning til Nynorsk: {$text}
Nynorsk: {$prediction}
\end{minted}

\noindent\footnotesize\texttt{\textbf{Prompt C:}}\vspace{-0.5em}
\begin{minted}[linenos=true, breaklines=true, baselinestretch=1.2, bgcolor=bg, breakanywhere=true, fontfamily=tt, fontsize=\footnotesize, xleftmargin=2em]{genshi}
Gje ei Nynorsk omsetjing av denne setninga: {$text}
Nynorsk: {$prediction}
\end{minted}

\newpage
\noindent\footnotesize\texttt{\textbf{Prompt D:}}\vspace{-0.5em}
\begin{minted}[linenos=true, breaklines=true, baselinestretch=1.2, bgcolor=bg, breakanywhere=true, fontfamily=tt, fontsize=\footnotesize, xleftmargin=2em]{genshi}
Kva blir "{$text}" på Nynorsk?
Nynorsk: {$prediction}
\end{minted}
}

\paragraph{Full results} The complete evaluation on translation to Nynorsk is given in \Cref{tab:en-se-results-by-shots-and-prompts} (BLEU scores in the top sub-table and chrF++ scores below).

\begin{table*}[!th]
\resizebox{\textwidth}{!}{%
\begin{tabular}{@{}l@{\hspace{1em}}*{14}{c}@{}}
\textsc{\textbf{BLEU scores}} \\
\toprule
& 
\multicolumn{4}{c}{\textbf{0-shot}}
& &
\multicolumn{4}{c}{\textbf{1-shot}}
& &
\multicolumn{4}{c}{\textbf{16-shot}}
\\
\textbf{Prompt template} & A & B & C & D & & A & B & C & D & & A & B & C & D \\
\midrule
NorMistral-11B & 36.2 & 6.9 & 20.1 & 39.3 & & 46.3\textsuperscript{\textpm1.6} & 45.4\textsuperscript{\textpm1.5} & 45.2\textsuperscript{\textpm1.4} & 44.7\textsuperscript{\textpm1.5} & & 46.5\textsuperscript{\textpm1.6} & \cellcolor{bestres}\textbf{48.0\textsuperscript{\textpm1.6}} & 46.1\textsuperscript{\textpm1.6} & \textbf{47.3\textsuperscript{\textpm1.6}} \\
\\[-0.5em]
NorwAI-Mistral-7B & \cellcolor{bestres}\textbf{46.0} & \textbf{44.7} & \textbf{42.3} & \textbf{40.0} & & \textbf{46.7\textsuperscript{\textpm1.6}} & \cellcolor{bestres}\textbf{46.7\textsuperscript{\textpm1.6}} & 46.6\textsuperscript{\textpm1.7} & \textbf{45.9\textsuperscript{\textpm1.6}} & & 47.4\textsuperscript{\textpm1.8} & 46.5\textsuperscript{\textpm1.8} & 46.1\textsuperscript{\textpm1.6} & 46.5\textsuperscript{\textpm1.7} \\
NorwAI-Llama2-7B & 43.9 & 23.2 & 0.0 & 28.8 & & 45.9\textsuperscript{\textpm1.7} & 46.7\textsuperscript{\textpm1.8} & \textbf{46.7\textsuperscript{\textpm1.7}} & 45.2\textsuperscript{\textpm1.7} & & \textbf{47.4\textsuperscript{\textpm1.7}} & 47.2\textsuperscript{\textpm1.7} & \textbf{47.3\textsuperscript{\textpm1.8}} & 47.0\textsuperscript{\textpm1.8} \\
NorMistral-7B-warm & 43.5 & 31.2 & 15.2 & 11.7 & & 43.7\textsuperscript{\textpm1.7} & 44.6\textsuperscript{\textpm1.8} & 43.5\textsuperscript{\textpm1.6} & 43.5\textsuperscript{\textpm1.8} & & 43.6\textsuperscript{\textpm1.8} & 44.7\textsuperscript{\textpm1.7} & 43.9\textsuperscript{\textpm1.7} & 44.2\textsuperscript{\textpm1.6} \\
NorGPT-3B & 1.5 & 2.4 & 0.8 & 1.6 & & 0.1\textsuperscript{\textpm0.0} & 0.2\textsuperscript{\textpm0.0} & 0.1\textsuperscript{\textpm0.0} & 0.4\textsuperscript{\textpm0.1} & & 0.2\textsuperscript{\textpm0.0} & 0.7\textsuperscript{\textpm0.1} & 2.6\textsuperscript{\textpm0.5} & 0.8\textsuperscript{\textpm0.1} \\
Viking-7B & 26.7 & 44.3 & 16.5 & 1.9 & & 45.0\textsuperscript{\textpm1.7} & 44.4\textsuperscript{\textpm1.6} & 43.7\textsuperscript{\textpm1.7} & 42.3\textsuperscript{\textpm1.6} & & 44.5\textsuperscript{\textpm1.6} & 43.9\textsuperscript{\textpm1.5} & 44.5\textsuperscript{\textpm1.6} & 45.6\textsuperscript{\textpm1.6} \\
Viking-13B & 42.5 & 31.6 & 11.1 & 1.7 & & 45.2\textsuperscript{\textpm1.7} & 45.2\textsuperscript{\textpm1.7} & 44.8\textsuperscript{\textpm1.7} & 42.4\textsuperscript{\textpm1.6} & & 45.2\textsuperscript{\textpm1.7} & 45.1\textsuperscript{\textpm1.6} & 45.5\textsuperscript{\textpm1.7} & 45.6\textsuperscript{\textpm1.7} \\
Mistral-Nemo-12B & 33.0 & 33.2 & 33.9 & 29.2 & & 33.6\textsuperscript{\textpm1.5} & 34.7\textsuperscript{\textpm1.4} & 33.9\textsuperscript{\textpm1.3} & 35.1\textsuperscript{\textpm1.5} & & 35.6\textsuperscript{\textpm1.6} & 35.4\textsuperscript{\textpm1.7} & 35.3\textsuperscript{\textpm1.7} & 35.7\textsuperscript{\textpm1.6} \\
\bottomrule \\

\textsc{\textbf{chrF++ scores}} \\
\toprule
& 
\multicolumn{4}{c}{\textbf{0-shot}}
& &
\multicolumn{4}{c}{\textbf{1-shot}}
& &
\multicolumn{4}{c}{\textbf{16-shot}}
\\
\textbf{Prompt template} & A & B & C & D & & A & B & C & D & & A & B & C & D \\
\midrule
NorMistral-11B & 62.1 & 32.4 & 53.3 & 62.2 & & 65.1\textsuperscript{\textpm1.1} & 64.6\textsuperscript{\textpm1.1} & 64.4\textsuperscript{\textpm1.0} & 63.9\textsuperscript{\textpm1.1} & & 65.2\textsuperscript{\textpm1.1} & \cellcolor{bestres}\textbf{66.5\textsuperscript{\textpm1.2}} & \textbf{65.7\textsuperscript{\textpm1.1}} & \textbf{65.9\textsuperscript{\textpm1.1}} \\
\\[-0.5em]
NorwAI-Mistral-7B & \cellcolor{bestres}\textbf{64.9} & 62.6 & \textbf{60.8} & \textbf{63.4} & & \cellcolor{bestres}\textbf{65.4\textsuperscript{\textpm1.1}} & \textbf{64.9\textsuperscript{\textpm1.1}} & 64.8\textsuperscript{\textpm1.1} & \textbf{64.5\textsuperscript{\textpm1.1}} & & \textbf{65.7\textsuperscript{\textpm1.2}} & 65.1\textsuperscript{\textpm1.2} & 65.0\textsuperscript{\textpm1.1} & 65.1\textsuperscript{\textpm1.1} \\
NorwAI-Llama2-7B & 63.5 & 41.9 & 3.4 & 45.7 & & 64.1\textsuperscript{\textpm1.2} & 64.8\textsuperscript{\textpm1.1} & \textbf{64.8\textsuperscript{\textpm1.1}} & 63.5\textsuperscript{\textpm1.1} & & \textbf{65.7\textsuperscript{\textpm1.2}} & 65.8\textsuperscript{\textpm1.1} & \textbf{65.7\textsuperscript{\textpm1.2}} & 65.2\textsuperscript{\textpm1.2} \\
NorMistral-7B-warm & 62.3 & 48.1 & 32.6 & 29.0 & & 62.7\textsuperscript{\textpm1.3} & 63.6\textsuperscript{\textpm1.3} & 63.0\textsuperscript{\textpm1.2} & 63.0\textsuperscript{\textpm1.3} & & 63.6\textsuperscript{\textpm1.2} & 64.6\textsuperscript{\textpm1.1} & 64.4\textsuperscript{\textpm1.1} & 63.9\textsuperscript{\textpm1.2} \\
NorGPT-3B & 15.5 & 16.6 & 12.2 & 16.9 & & 4.9\textsuperscript{\textpm0.2} & 4.4\textsuperscript{\textpm0.2} & 3.3\textsuperscript{\textpm0.2} & 7.7\textsuperscript{\textpm0.5} & & 4.0\textsuperscript{\textpm0.2} & 9.2\textsuperscript{\textpm0.5} & 15.0\textsuperscript{\textpm1.1} & 8.0\textsuperscript{\textpm0.4} \\
Viking-7B & 56.9 & \textbf{63.6} & 47.8 & 15.1 & & 64.1\textsuperscript{\textpm1.2} & 64.4\textsuperscript{\textpm1.1} & 64.6\textsuperscript{\textpm1.1} & 62.8\textsuperscript{\textpm1.2} & & 64.5\textsuperscript{\textpm1.1} & 64.3\textsuperscript{\textpm1.0} & 64.5\textsuperscript{\textpm1.1} & 65.2\textsuperscript{\textpm1.1} \\
Viking-13B & 62.9 & 60.4 & 44.0 & 14.9 & & 64.7\textsuperscript{\textpm1.1} & 64.8\textsuperscript{\textpm1.1} & 64.4\textsuperscript{\textpm1.1} & 61.9\textsuperscript{\textpm1.2} & & 64.6\textsuperscript{\textpm1.1} & 64.9\textsuperscript{\textpm1.1} & 65.0\textsuperscript{\textpm1.1} & 65.3\textsuperscript{\textpm1.1} \\
Mistral-Nemo-12B & 55.4 & 54.2 & 55.5 & 53.8 & & 56.3\textsuperscript{\textpm1.1} & 57.0\textsuperscript{\textpm1.1} & 56.4\textsuperscript{\textpm1.0} & 56.9\textsuperscript{\textpm1.1} & & 57.3\textsuperscript{\textpm1.2} & 57.2\textsuperscript{\textpm1.2} & 57.2\textsuperscript{\textpm1.2} & 57.5\textsuperscript{\textpm1.2} \\
\bottomrule 
\end{tabular}
}
\caption{\textbf{Complete results on translation from English to Nynorsk}\hspace{1em}We show the detailed results for each evaluated model, few-shot setting and prompt template. As the few-shot demonstrations are sampled randomly, we repeat them five times and show the mean accuracy as well as the standard deviation (rendered as superscript). The best results for each column are boldfaced, the overall best result for each few-shot setting is highlighted in blue.}
\end{table*}

\subsubsection{English to Northern Sámi translation}

We source the data from the English-Sámi parallel corpus from Tatoeba \citep{tiedemann-2020-tatoeba}, specifically the latest \texttt{v2023-04-12} revision avaiable on HuggingFace at {\footnotesize\textbf{\url{https://hf.co/datasets/Helsinki-NLP/tatoeba}}}. We deduplicate this corpus (both on the source and target side) and remove the empty entries -- obtaining 53 examples in total. 

\paragraph{Prompt templates} We used the following five prompt templates for testing all language models on translation to Northern Sámi.

{
\vspace{1em}
\noindent\footnotesize\texttt{\textbf{Prompt A:}}\vspace{-0.5em}
\begin{minted}[linenos=true, breaklines=true, baselinestretch=1.2, bgcolor=bg, breakanywhere=true, fontfamily=tt, fontsize=\footnotesize, xleftmargin=2em]{genshi}
Eaŋgalsgiella: {$text}
Davvisámegiella: {$prediction}
\end{minted}

\noindent\footnotesize\texttt{\textbf{Prompt B:}}\vspace{-0.5em}
\begin{minted}[linenos=true, breaklines=true, baselinestretch=1.2, bgcolor=bg, breakanywhere=true, fontfamily=tt, fontsize=\footnotesize, xleftmargin=2em]{genshi}
Engelsk: {$text}
Samisk: {$prediction}
\end{minted}

\noindent\footnotesize\texttt{\textbf{Prompt C:}}\vspace{-0.5em}
\begin{minted}[linenos=true, breaklines=true, baselinestretch=1.2, bgcolor=bg, breakanywhere=true, fontfamily=tt, fontsize=\footnotesize, xleftmargin=2em]{genshi}
Oversett følgende setning til nordsamisk: {$text}
Nordsamisk: {$prediction}
\end{minted}

\noindent\footnotesize\texttt{\textbf{Prompt D:}}\vspace{-0.5em}
\begin{minted}[linenos=true, breaklines=true, baselinestretch=1.2, bgcolor=bg, breakanywhere=true, fontfamily=tt, fontsize=\footnotesize, xleftmargin=2em]{genshi}
Gi en oversettelse til nordsamisk for denne setningen: {$text}
Nordsamisk: {$prediction}
\end{minted}

\noindent\footnotesize\texttt{\textbf{Prompt E:}}\vspace{-0.5em}
\begin{minted}[linenos=true, breaklines=true, baselinestretch=1.2, bgcolor=bg, breakanywhere=true, fontfamily=tt, fontsize=\footnotesize, xleftmargin=2em]{genshi}
Hva blir "{$text}" på nordsamisk?
Nordsamisk: {$prediction}
\end{minted}
}

\paragraph{Full results} The complete evaluation on translation to Sámi is given in \Cref{tab:en-se-results-by-shots-and-prompts} (BLEU scores in the top sub-table and chrF++ scores below).

\begin{table*}[!th]
\resizebox{\textwidth}{!}{%
\begin{tabular}{@{}l@{\hspace{1em}}*{17}{c}@{}}
\textsc{\textbf{BLEU scores}} \\
\toprule
& 
\multicolumn{5}{c}{\textbf{0-shot}} & &
\multicolumn{5}{c}{\textbf{1-shot}} & & 
\multicolumn{5}{c}{\textbf{16-shot}} \\
\textbf{Prompt template} & A & B & C & D & E & & A & B & C & D & E & & A & B & C & D & E \\
\midrule
NorMistral-11B & \cellcolor{bestres}\textbf{31.5} & \textbf{31.0} & 14.8 & \textbf{23.2} & 14.7 & & 24.8\textsuperscript{\textpm5.1} & \cellcolor{bestres}\textbf{44.1\textsuperscript{\textpm1.2}} & 15.7\textsuperscript{\textpm1.6} & 20.2\textsuperscript{\textpm9.4} & \textbf{33.7\textsuperscript{\textpm8.2}} & & \textbf{45.5\textsuperscript{\textpm2.2}} & \textbf{48.8\textsuperscript{\textpm2.4}} & \textbf{49.0\textsuperscript{\textpm2.1}} & \textbf{48.8\textsuperscript{\textpm2.3}} & \cellcolor{bestres}\textbf{50.4\textsuperscript{\textpm0.9}} \\
\\[-0.5em]
NorwAI-Mistral-7B & 21.2 & 25.0 & \textbf{24.0} & 16.7 & \textbf{20.7} & & \textbf{26.6\textsuperscript{\textpm1.7}} & 24.7\textsuperscript{\textpm2.2} & 24.4\textsuperscript{\textpm1.6} & \textbf{24.2\textsuperscript{\textpm2.4}} & 25.0\textsuperscript{\textpm1.1} & & 24.8\textsuperscript{\textpm2.7} & 25.1\textsuperscript{\textpm2.6} & 26.2\textsuperscript{\textpm1.8} & 26.8\textsuperscript{\textpm2.0} & 27.5\textsuperscript{\textpm2.3} \\
NorwAI-Llama2-7B & 16.9 & 10.7 & 3.3 & 0.0 & 15.4 & & 24.7\textsuperscript{\textpm2.1} & 23.5\textsuperscript{\textpm2.7} & \textbf{24.5\textsuperscript{\textpm2.1}} & 22.6\textsuperscript{\textpm2.5} & 22.5\textsuperscript{\textpm1.8} & & 26.6\textsuperscript{\textpm2.3} & 27.6\textsuperscript{\textpm2.0} & 27.9\textsuperscript{\textpm1.9} & 27.6\textsuperscript{\textpm2.1} & 28.5\textsuperscript{\textpm1.6} \\
NorMistral-7B-warm & 12.2 & 5.7 & 0.0 & 0.0 & 0.0 & & 11.2\textsuperscript{\textpm3.1} & 10.5\textsuperscript{\textpm1.9} & 12.2\textsuperscript{\textpm3.2} & 10.5\textsuperscript{\textpm2.0} & 8.4\textsuperscript{\textpm0.5} & & 14.9\textsuperscript{\textpm2.2} & 17.2\textsuperscript{\textpm1.9} & 17.9\textsuperscript{\textpm1.5} & 15.9\textsuperscript{\textpm1.2} & 18.5\textsuperscript{\textpm2.5} \\
NorGPT-3B & 0.0 & 0.0 & 0.0 & 0.0 & 0.0 & & 0.0\textsuperscript{\textpm0.0} & 0.0\textsuperscript{\textpm0.0} & 0.0\textsuperscript{\textpm0.0} & 0.0\textsuperscript{\textpm0.0} & 0.0\textsuperscript{\textpm0.0} & & 0.0\textsuperscript{\textpm0.0} & 0.0\textsuperscript{\textpm0.0} & 0.0\textsuperscript{\textpm0.0} & 0.0\textsuperscript{\textpm0.0} & 0.0\textsuperscript{\textpm0.0} \\
Viking-7B & 0.0 & 0.0 & 0.0 & 0.0 & 0.0 & & 0.0\textsuperscript{\textpm0.0} & 1.0\textsuperscript{\textpm2.3} & 0.7\textsuperscript{\textpm1.5} & 0.7\textsuperscript{\textpm1.6} & 0.5\textsuperscript{\textpm1.1} & & 4.6\textsuperscript{\textpm2.6} & 6.8\textsuperscript{\textpm1.8} & 6.7\textsuperscript{\textpm2.6} & 7.8\textsuperscript{\textpm1.6} & 4.3\textsuperscript{\textpm3.9} \\
Viking-13B & 0.0 & 0.0 & 0.0 & 0.0 & 0.0 & & 0.3\textsuperscript{\textpm0.7} & 3.3\textsuperscript{\textpm3.2} & 2.3\textsuperscript{\textpm2.1} & 4.3\textsuperscript{\textpm4.0} & 1.8\textsuperscript{\textpm1.9} & & 9.8\textsuperscript{\textpm1.1} & 11.6\textsuperscript{\textpm1.9} & 11.5\textsuperscript{\textpm1.9} & 11.4\textsuperscript{\textpm2.0} & 11.2\textsuperscript{\textpm1.2} \\
Mistral-Nemo-12B & 0.0 & 0.0 & 0.0 & 0.0 & 0.0 & & 0.0\textsuperscript{\textpm0.0} & 1.4\textsuperscript{\textpm2.0} & 0.8\textsuperscript{\textpm1.8} & 0.9\textsuperscript{\textpm2.0} & 0.9\textsuperscript{\textpm1.9} & & 3.9\textsuperscript{\textpm2.5} & 6.5\textsuperscript{\textpm2.2} & 4.9\textsuperscript{\textpm3.2} & 5.7\textsuperscript{\textpm3.8} & 6.1\textsuperscript{\textpm1.4} \\
\bottomrule\\

\textsc{\textbf{chrF++ scores}} \\
\toprule
& 
\multicolumn{5}{c}{\textbf{0-shot}} & &
\multicolumn{5}{c}{\textbf{1-shot}} & & 
\multicolumn{5}{c}{\textbf{16-shot}} \\
\textbf{Prompt template} & A & B & C & D & E & & A & B & C & D & E & & A & B & C & D & E \\
\midrule
NorMistral-11B & \textbf{51.8} & \cellcolor{bestres}\textbf{58.8} & 49.5 & \textbf{58.2} & 47.3 & & \textbf{55.5\textsuperscript{\textpm2.4}} & \cellcolor{bestres}\textbf{66.7\textsuperscript{\textpm0.8}} & \textbf{54.6\textsuperscript{\textpm1.8}} & \textbf{56.9\textsuperscript{\textpm5.4}} & \textbf{63.2\textsuperscript{\textpm3.7}} & & \textbf{64.9\textsuperscript{\textpm1.2}} & \textbf{67.2\textsuperscript{\textpm0.6}} & \textbf{67.9\textsuperscript{\textpm1.0}} & \textbf{67.7\textsuperscript{\textpm1.4}} & \cellcolor{bestres}\textbf{69.6\textsuperscript{\textpm0.9}} \\
\\[-0.5em]
NorwAI-Mistral-7B & 44.9 & 45.8 & \textbf{53.9} & 50.6 & \textbf{49.6} & & 50.7\textsuperscript{\textpm1.3} & 50.8\textsuperscript{\textpm1.5} & 50.8\textsuperscript{\textpm0.8} & 51.5\textsuperscript{\textpm1.1} & 50.4\textsuperscript{\textpm0.7} & & 51.0\textsuperscript{\textpm1.7} & 51.7\textsuperscript{\textpm1.7} & 52.3\textsuperscript{\textpm1.7} & 52.4\textsuperscript{\textpm1.2} & 52.8\textsuperscript{\textpm1.6} \\
NorwAI-Llama2-7B & 37.3 & 30.0 & 46.8 & 44.1 & 39.3 & & 45.6\textsuperscript{\textpm2.0} & 44.2\textsuperscript{\textpm3.4} & 48.1\textsuperscript{\textpm2.3} & 46.6\textsuperscript{\textpm1.7} & 44.0\textsuperscript{\textpm1.4} & & 48.4\textsuperscript{\textpm1.9} & 50.2\textsuperscript{\textpm1.3} & 50.4\textsuperscript{\textpm1.6} & 50.2\textsuperscript{\textpm1.4} & 50.4\textsuperscript{\textpm1.1} \\
NorMistral-7B-warm & 27.4 & 22.5 & 21.4 & 26.8 & 23.8 & & 32.3\textsuperscript{\textpm2.4} & 30.9\textsuperscript{\textpm1.9} & 32.7\textsuperscript{\textpm2.2} & 32.7\textsuperscript{\textpm1.5} & 29.9\textsuperscript{\textpm0.7} & & 38.1\textsuperscript{\textpm3.7} & 37.7\textsuperscript{\textpm3.5} & 38.9\textsuperscript{\textpm1.5} & 37.2\textsuperscript{\textpm1.5} & 39.4\textsuperscript{\textpm2.1} \\
NorGPT-3B & 2.7 & 3.0 & 3.0 & 2.8 & 2.8 & & 2.8\textsuperscript{\textpm0.1} & 2.8\textsuperscript{\textpm0.1} & 2.8\textsuperscript{\textpm0.1} & 2.8\textsuperscript{\textpm0.1} & 2.9\textsuperscript{\textpm0.1} & & 2.9\textsuperscript{\textpm0.1} & 2.6\textsuperscript{\textpm0.1} & 3.0\textsuperscript{\textpm0.1} & 3.0\textsuperscript{\textpm0.1} & 3.1\textsuperscript{\textpm0.1} \\
Viking-7B & 10.4 & 11.5 & 11.5 & 11.7 & 8.1 & & 10.3\textsuperscript{\textpm0.9} & 13.0\textsuperscript{\textpm1.7} & 12.3\textsuperscript{\textpm1.5} & 12.7\textsuperscript{\textpm1.1} & 12.4\textsuperscript{\textpm1.3} & & 18.0\textsuperscript{\textpm1.3} & 19.6\textsuperscript{\textpm0.7} & 19.1\textsuperscript{\textpm1.3} & 19.7\textsuperscript{\textpm1.0} & 19.5\textsuperscript{\textpm1.6} \\
Viking-13B & 11.3 & 10.8 & 8.0 & 10.6 & 6.2 & & 10.9\textsuperscript{\textpm0.7} & 14.9\textsuperscript{\textpm1.4} & 16.2\textsuperscript{\textpm0.8} & 16.1\textsuperscript{\textpm0.7} & 15.2\textsuperscript{\textpm1.5} & & 24.5\textsuperscript{\textpm2.5} & 24.5\textsuperscript{\textpm2.0} & 24.9\textsuperscript{\textpm2.2} & 24.0\textsuperscript{\textpm2.9} & 25.2\textsuperscript{\textpm1.7} \\
Mistral-Nemo-12B & 11.5 & 12.1 & 12.2 & 12.4 & 10.5 & & 14.2\textsuperscript{\textpm2.0} & 14.2\textsuperscript{\textpm0.9} & 15.3\textsuperscript{\textpm1.1} & 14.8\textsuperscript{\textpm1.3} & 14.9\textsuperscript{\textpm0.9} & & 21.9\textsuperscript{\textpm0.8} & 22.3\textsuperscript{\textpm1.7} & 22.6\textsuperscript{\textpm1.5} & 22.4\textsuperscript{\textpm1.6} & 23.0\textsuperscript{\textpm1.5} \\
\bottomrule
\end{tabular}
}
\caption{\textbf{Complete results on translation from English to Northern Sámi}\hspace{1em}We show the detailed results for each evaluated model, few-shot setting and prompt template. As the few-shot demonstrations are sampled randomly, we repeat them five times and show the mean accuracy as well as the standard deviation (rendered as superscript). The best results for each column are boldfaced, the overall best result for each few-shot setting is highlighted in blue.}

\label{tab:en-se-results-by-shots-and-prompts}
\end{table*}

\end{document}